\documentclass[letterpaper, 10 pt, journal, twoside]{IEEEtran}
\usepackage[T1]{fontenc}
\usepackage[latin9]{inputenc}
\usepackage{geometry}
\usepackage{color}
\usepackage{float}
\usepackage{textcomp}
\usepackage{amsmath}
\usepackage{amsthm}
\usepackage{amssymb}
\usepackage{stackrel}
\usepackage{graphicx}
\usepackage{hyperref}

\makeatletter

\newcommand{\lyxmathsym}[1]{\ifmmode\begingroup\def\b@ld{bold}
  \text{\ifx\math@version\b@ld\bfseries\fi#1}\endgroup\else#1\fi}

\floatstyle{ruled}
\newfloat{algorithm}{tbp}{loa}
\providecommand{\algorithmname}{Algorithm}
\floatname{algorithm}{\protect\algorithmname}

\providecommand{\tabularnewline}{\\}
\floatstyle{ruled}
\newfloat{algorithm}{tbp}{loa}
\providecommand{\algorithmname}{Algorithm}
\floatname{algorithm}{\protect\algorithmname}

\theoremstyle{definition}
\newtheorem{example}{\protect\examplename}
\theoremstyle{definition}
\newtheorem{problem}{\protect\problemname}
\theoremstyle{definition}
\newtheorem{defn}{\protect\definitionname}
\theoremstyle{plain}

\theoremstyle{plain}

\theoremstyle{plain}

\usepackage{cite}
\usepackage{algpseudocode}
\usepackage{setspace}
\IEEEoverridecommandlockouts

\makeatother

\usepackage{babel}
\providecommand{\corollaryname}{Corollary}
\providecommand{\definitionname}{Definition}
\providecommand{\examplename}{Example}
\providecommand{\lemmaname}{Lemma}
\providecommand{\problemname}{Problem}
\providecommand{\theoremname}{Theorem}

\begin{document}
\title{A Novel Task-Driven {Diffusion-Based Policy with Affordance Learning} for Generalizable Manipulation of Articulated
Objects\thanks{Hao Zhang, Zhen Kan (Corresponding Author), and Weiwei Shang
are with the Department of Automation at the University of Science
and Technology of China, Hefei, Anhui, China. Yongduan Song is with the School of Automation, Chongqing University, Chongqing, China.} \thanks{This work was supported in part by National Key R\&D Program of China under Grant 2022YFB4701400/4701403 and National Natural Science Foundation of China under Grant U201360.}}
\author{Hao Zhang, Zhen Kan, Weiwei Shang, and Yongduan Song}
\maketitle
\begin{abstract}
Despite recent advances in dexterous manipulations, the manipulation of articulated objects and generalization across different categories remain significant challenges. To address these issues, we introduce DART, a novel framework that enhances a {\underline{d}}{iffusion-based
policy} with \underline{a}ffo\underline{r}dance learning and linear \underline{t}emporal logic (LTL) representations to improve the learning efficiency and generalizability of articulated dexterous manipulation. Specifically, DART leverages LTL to understand task semantics and affordance learning to identify optimal interaction points. The {diffusion-based policy} then generalizes these interactions across various categories. Additionally, we exploit an optimization method based on interaction data to refine actions, overcoming the limitations of traditional diffusion policies that typically rely on offline reinforcement learning or learning from demonstrations. Experimental results demonstrate that DART outperforms most existing methods in manipulation ability, generalization performance, transfer reasoning, and robustness. For more information, visit our project website at: https://sites.google.com/view/dart0257/.

\global\long\def\prog{\operatorname{prog}}%
\global\long\def\argmax{\operatorname{argmax}}%
\global\long\def\argmin{\operatorname{argmin}}%
\end{abstract}

\section{Introduction\label{sec:Intro}}

The manipulation of articulated objects has been an interesting
and important topic in robotic learning. Although prior research has demonstrated promising results in the manipulation of rigid bodies, significant challenges persist when it comes to handling articulated objects 
\cite{wang2022learning}. {Generalizing to various types of articulated objects \cite{xu2022universal} is particularly
difficult for dexterous manipulations.} For example, if a dexterous
hand can open the lid of a toilet, it should also be capable of opening the lid of a garbage can, despite their cosmetic differences. While many recent efforts have focused on
improving the robotic generalization performance \cite{lu2021constrained}
or reducing the exploration burden \cite{wang2023task}, enhancing the learning efficiency or improve the generalization
ability for high degrees of freedom (DOF) skills, such as dexterous manipulation, remains a challenging problem, not to mention achieving both simultaneously.

There are various methods enabling skill generalization via reinforcement learning (RL).
Taking advantage of Meta-RL to improve
the generalization potential of algorithms is an effective approach 
\cite{Rakelly2019}. However,
solving context Markov decision processes (MDPs) in Meta-RL often involves cumbersome computation.
Recently, increasing efforts are being dedicated to leveraging vision features to enhance RL in tackling various challenging tasks \cite{yarats2021mastering,nair2022r3m,gmelin2023efficient}.
One of the interesting findings is that algorithms exploiting point
cloud information can have good generalization without any Meta-RL-like
training \cite{bao2023dexart}. However,
most of the previous work struggles to understand the geometry of articulated objects, thus making it challenging to identify appropriate interaction points when developing manipulation skills.

One of the effective ways to address the above shortcomings is affordance
learning \cite{gibson1977theory}.
In \cite{mo2021where2act}, a learning-from-interaction framework
was proposed to manipulate articulated objects, which uses an actionability scoring module to estimate the score of each pixel and provide better interaction positions. The work of \cite{wu2023learning} further extends
\cite{mo2021where2act} by exploiting
the multi-stage stable learning and self-supervised data collection
to achieve deformable object manipulation. In \cite{10161571},  RL was guided by affordance learning to achieve diverse
manipulation, which yields an end-to-end learning framework to avoid offline data collection. Although
affordance learning has shown great potential in the
manipulation of articulated objects, generating a high-quality affordance map often requires a significant amount of interactions for training, which significantly impacts learning efficiency in long-horizon tasks.

Due to the rich expressivity, linear temporal logic
(LTL) \cite{Baier2008} is capable of describing long-horizon tasks composed of logically organized sub-tasks.
For instance, LTL specification was converted to a finite-state
predicate automaton in \cite{Li2019} to facilitate the use of RL in
performing complex manipulation tasks. However, previous methods mainly focus on solving tasks with different
automaton, without considering the task semantics of LTL instructions. Based on graph neural networks, an approach exploiting the compositional
syntax and semantics of LTL was
proposed in \cite{vaezipoor2021ltl2action} to learn task-conditioned policies, which is generalizable to new instructions. In \cite{zhang2023exploiting},
Transformer \cite{vaswani2017attention} was employed to encode the
LTL specifications for improved representation capability and interpretability. {Direct language approaches
that utilize large language models \cite{wang2024large}
or represent tasks using text \cite{jung2024touch}
have brought success to manipulation at the task level. However, these
approaches are limited either by their expressiveness and rigor, or
by the computational efficiency of the algorithms relative to LTL.} Despite recent progress, neither vision-based RL nor LTL-guided motion planning inherently improves the generalization
of the learned manipulation skills.

As an effective generative model, the diffusion model (DM) \cite{sohl2015deep,ho2020denoising,song2020score} utilizes a parametric reverse
process to generate samples from a Gaussian distribution.
Inspired by this, DM has been investigated for long-horizon manipulation  \cite{chi2023diffusion,yang2023policy,shi2024superresolution}. However, in most of these studies, DM has functioned as a policy within either an offline reinforcement learning or imitation learning framework. Extending this approach to dexterous manipulations is challenging due to the difficulty in obtaining high-quality operational data. Our primary motivation is to determine whether task semantics can be leveraged to drive affordance learning and guide the agent in achieving {a policy with the
diffusion model}, thereby improving learning efficiency for articulated objects and enhancing generalization in operational performance. To address this, we propose DART, a novel policy generalization framework that enhances {a} \textbf{{d}}{iffusion-based policy} using \textbf{a}ffo\textbf{r}dance learning and linear \textbf{t}emporal logic representations to improve both generalizability and learning efficiency. DART guides the agent in improving sampling efficiency by encoding task semantics into LTL representations.  Additionally, it utilizes affordance learning to generate an affordance map based on point cloud information captured by the camera, predicting optimal contact points for dexterous manipulation. Furthermore, DART incorporates a task-driven action optimization method for the {policy}, enabling training and optimization in online settings. 

\textbf{Contributions:} The main contributions of this work are summarized
as follows:
\begin{itemize}
\item We introduce a novel policy generalization algorithm that enhances {a diffusion-based policy} through the use of LTL representations and affordance learning, thereby improving learning efficiency and ensuring robust generalization performance.
\item We construct a task-driven MDP based on LTL representations, which not only incorporates task semantics to facilitate robotic learning but also designs a contact planner based on affordance learning to enhance the efficiency of articulated manipulations.
\item {We exploit a diffusion-based policy for interacting
with articulated objects, whose multimodal policy representation lays
the groundwork for the generalization.}
\item {We further extend the task environment of \cite{bao2023dexart} and evaluate the methods when generalizing to different
category-level objects. A comprehensive empirical study validates
that DART significantly outperforms baseline methods in terms of manipulation
ability, generalization performance, transfer reasoning, and robustness.}
\end{itemize}

\section{Preliminaries and Problem Formulation\label{sec:Pre}}

\subsection{sc-LTL and LTL Progression\label{subsec:LTL Progression}}

Co-safe LTL (sc-LTL) is a subclass of LTL that can be satisfied by
finite-horizon state trajectories \cite{Kupferman2001}. Since sc-LTL
is suitable for describing robotic tasks, this work focuses
on sc-LTL. An sc-LTL formula is built on a set of atomic propositions
$\Pi$ that can be true or false, standard Boolean operators such
as $\wedge$ (conjunction), $\lor$ (disjunction), and $\lnot$ (negation),
temporal operators such as $\bigcirc$ (next), $\diamondsuit$ (eventually),
and $\cup$ (until). The semantics of an sc-LTL formula are interpreted
over a word $\boldsymbol{\sigma}=\sigma_{0}\sigma_{1}...\sigma_{n}$,
which is a finite sequence with $\sigma_{i}\in2^{\Pi}$, $i=0,\ldots,n$,
where $2^{\Pi}$ represents the power set of $\Pi$. Denote by $\left\langle \boldsymbol{\sigma},i\right\rangle \vDash\varphi$
if the sc-LTL formula $\varphi$ holds from position $i$ of $\boldsymbol{\sigma}$.

LTL formulas can also be progressed along a sequence of truth assignment
\cite{tuli2022learning}. Specifically,
give an LTL formula $\varphi$ and a word $\boldsymbol{\sigma}=\sigma_{0}\sigma_{1}...$, 
the LTL progression $\prog\left(\sigma_{i},\varphi\right)$ at step $i$ is $\prog\left(\sigma_{i},p\right)=\mathrm{True}\text{ if }p\in\sigma_{i}\text{, where }p\in\Pi$
and $\prog\left(\sigma_{i},p\right)=\mathrm{False}$ otherwise. The
operator $\mathrm{prog}$ takes an LTL formula $\varphi$ and the
current label $\sigma_{i}$ as input at each step, and outputs a formula
indicating the remaining instructions to be addressed.

\subsection{Labeled MDP and Visual Reinforcement Learning}

When performing the sc-LTL task $\varphi$ in visual RL, the interaction
between the robot and the environment can be modeled by a labeled
MDP $\mathcal{M}_{e}=\left(S,T,A,p_{e},\Pi,L,R,\gamma,\mu,\Omega,O\right)$,
where $S$ is the state space, $T\subseteq S$ is a set of terminal
states, $A$ is the action space, $p_{e}(s'|s,a)$ is the transition
probability from $s\in S$ to $s'\in S$ under action $a\in A$, $\Pi$
is a set of atomic propositions indicating the properties associated
with the states, $L:S\rightarrow2^{\Pi}$ is the labeling function,
$R:S\rightarrow\mathbb{R}$ is the reward function, $\gamma\in\left(0,1\right]$
is the discount factor, $\mu$ is the initial state distribution,
$\Omega$ is the observation space, and $O:S\rightarrow\Omega$
is the observation function that maps an environment state $s$ to
an observation in $\Omega$. The labeling function
$L$ can be seen as a set of event detectors that trigger when $p\in\Pi$
presents in the environment, allowing the robot to determine whether
or not an LTL specification is satisfied. 

For any task $\varphi$, the robot interacts with the environment
following a deterministic policy $\pi_{e}$ over $\mathcal{M}_{e}$, i.e., given a state $s_{t}$ at time $t$, $\pi_{e}:\Omega\rightarrow A$ maps an observation $O(s_{t})$
to an action $a \in A$. The robot then receives
a reward $r_{t}=R(s_{t})$. Given a task $\varphi$, the goal of
the agent is to learn an optimal policy $\pi^{*}(a|o)$ that maximizes
the expected discounted return $E\left[\stackrel[k=0]{\infty}{\sum}\gamma^{k}r_{t+k}\mid O_{t}=o\right]$
given an observation $o\in O$.

\subsection{SDE in Diffusion Model \label{subsec:DM}}

{Diffusion model \cite{sohl2015deep,ho2020denoising,song2020score},
as a generative model, transforms any data distribution into a simple
Gaussian distribution by adding noise (forward process) and then denoise
it using neural networks (reverse process) with parameterized estimators.} The forward noising process
in continuous time, denoted as $dx_{t}$, add noise to the data $x\in\mathbb{R}^{p}$ following the Stochastic Differential
Equation (SDE) 
\begin{equation}
dx_{t}=\mathbf{f}\left(x_{t},t\right)dt+g\left(t\right)d\mathbf{w}_{t}\label{eq: forward in DM}
\end{equation}
where $\mathbf{f}\left(\cdot,t\right):\mathbb{R}^{p}\rightarrow\mathbb{R}^{p}$ is a Lipschitz function, $\mathbf{w}$
is a standard Brownian motion and $g\left(t\right)\in\mathbb{R}$ is a noise
schedule which injects a fixed series of noise levels monotonically
increasing via forward diffusion $t$. The reverse denoising process,
denoted as $d\bar{x}_{t}$, generates new samples by utilizing the
results of \cite{song2020score} for sampling 
\begin{equation}
d\bar{x}_{t}=\left[\mathbf{f}\left(\bar{x}_{t},t\right)-g^{2}\left(t\right)\nabla\log p_{t}\left(\bar{x}_{t}\right)\right]dt+g\left(t\right)d\mathbf{\bar{w}}_{t}\label{eq:reverse in DM}
\end{equation}
 where $\nabla\log p_{t}\left(\cdot\right)$ represents the gradient of the logarithm of the marginal
density function of the data $x_{t}$ and $\mathbf{\bar{w}}$
denotes a reverse Brownian motion. It is proposed in \cite{ho2020denoising}
to utilize the noise estimator $\varepsilon_{\xi}\left(\bar{x}_{t},t\right)$
as the score of distribution of $\bar{x}_{t}$ to approximate $\nabla\log p_{t}\left(\bar{x}_{t}\right)$
so that (\ref{eq:reverse in DM}) can be rewritten as 
\[
d\bar{x}_{t}=\left[\mathbf{f}\left(\bar{x}_{t},t\right)-g^{2}\left(t\right)\varepsilon_{\xi}\left(\bar{x}_{t},t\right)\right]dt+g\left(t\right)d\mathbf{\bar{w}}_{t}.
\]
In this work, we will replace the above data $x$ with action $a$
to design the decision-making process in RL using the diffusion-based policy
for better generalization to different tasks.

\subsection{Challenges and Problem Formulation}

\begin{figure*}
\centering{}\includegraphics[scale=0.28]{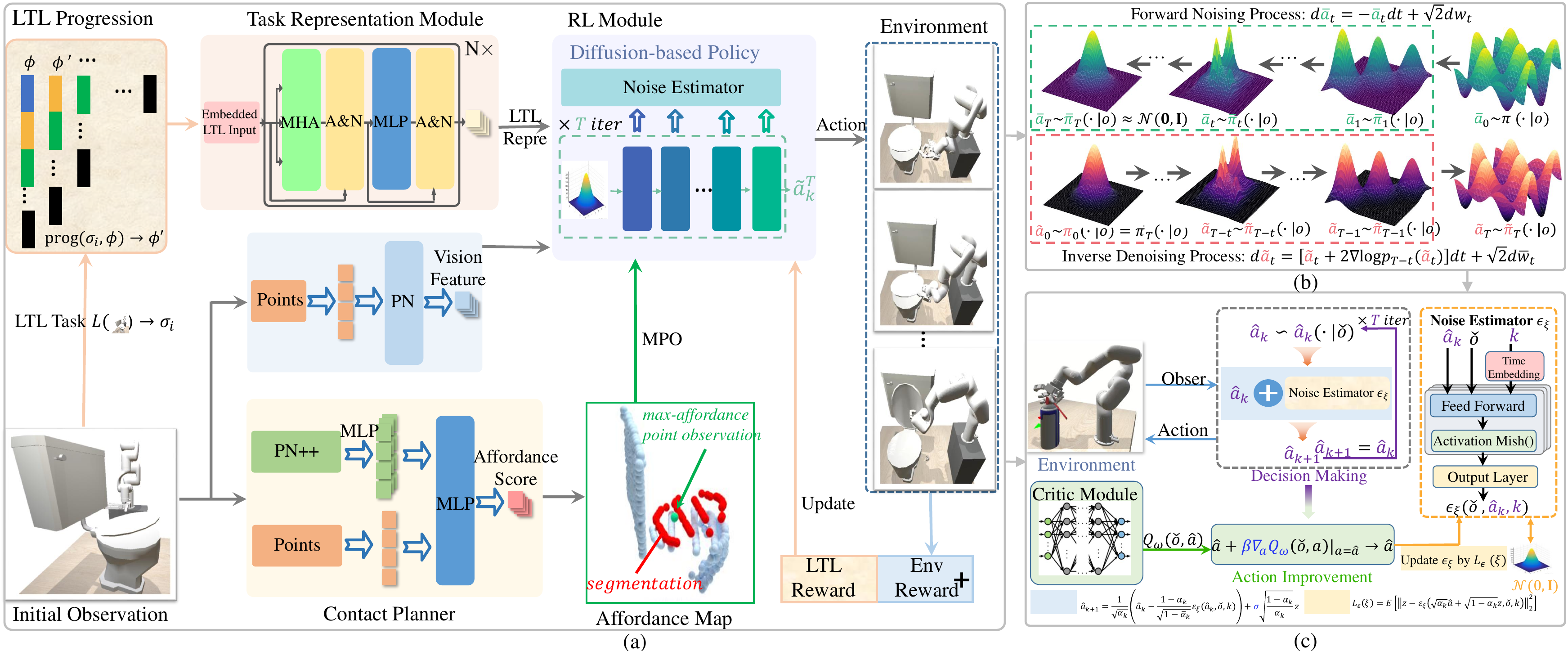}\caption{\label{fig:Framework of DART}(a) The framework of DART which consists
of three key modules. Task representation module: the Transformer encoder
architecture is employed to encode LTL instruction $\varphi$ to facilitate the understanding of task semantics. Contact Planner: affordance
learning is applied to generate an affordance map based on observation
and predict the MPO. RL Module: {A diffusion-based
policy} is used
to interact with the objects and improve the generalization performance. (b) The outlines of {a
series of forward noise addition policy distributions
and a series of inverse denoising policy distributions. (c) The outline of } a {diffusion-based}
policy and action improvement.}
\end{figure*}

To elaborate the problem and challenges, the following running example will be used throughout the work. 
\begin{example}
\label{example1}Consider a dexterous manipulation \cite{bao2023dexart}
as shown in Fig. \ref{fig:Framework of DART}(a), in which the robot needs
to approach the toilet and open the lid $O_{lid}$. The set of propositions
$\Pi$ is \{$\mathsf{toilet\_approached}$, $\mathsf{lid\_grasped}$,
$\mathsf{lid\_opened}$\}. Using above propositions in $\Pi$, an
example sc-LTL formula is $\varphi_{\mathsf{\mathsf{toilet}}}=\lozenge({\mathsf{toilet\_approached}}\wedge\lozenge(\mathsf{lid\_grasped}\wedge\lozenge\mathsf{lid\_opened}))$,
which requires the robot to sequentially approach, grasp, and open the
lid of the toilet.
 
\end{example}

There are three main challenges in Example \ref{example1}. The first
challenge is how to understand the task semantics and thus improve the learning efficiency. The second challenge is how to predict the optimal contact points for dexterous manipulation of articulated
objects based on point cloud information. The third challenge is how to generalize the learned manipulation to different objects of the same kind, or even different category-level
objects. To this end, we aim to develop a trainable and generalizable policy $\pi_{\mathcal{\varphi}}$
that can overcome these challenges. Hence, the problem can be presented as follows.
\begin{problem}
\label{Prob1}Given a labeled MDP $\mathcal{M}_{e}=\left(S,T,A,p_{e},\Pi,L,R,\gamma,\mu,\Omega,O\right)$
corresponding to task $\varphi$ with the reward function $R_{\varphi}(s_{0}s_{1}...s_{t})$
to be designed, the goal of this work is to find an optimal policy
$\pi_{\mathcal{\varphi}}^{*}$ over the LTL instruction $\varphi$,
so that the successful rate evaluated on seen and unseen sets of articulated
objects under the policy $\pi_{\mathcal{\varphi}}^{*}$ can be maximized.
\end{problem}

\section{Manipulation Generalization\label{sec:Algorithm Design}}

In this section, we present a novel framework, namely task-driven {diffusion-based policy} with affordance learning for generalizable manipulation of articulated objects (DART),  that enables the agent to learn efficiently and generalize diversely. We first
describe the overview of DART in Sec. \ref{subsec:Overview} and
then explain the technical details in Sec. \ref{subsec:Segmentation}
to Sec. \ref{subsec:Diffusion Policy}.

\subsection{Overview of DART\label{subsec:Overview}}

As shown in Fig. \ref{fig:Framework of DART},
DART uses LTL representations and affordance learning to guide the 
{diffusion-based policy} to improve the learning efficiency and the articulated
dexterous generalization. Since traditional diffusion policies highly rely on demonstrations, a task-oriented
optimization approach is incorporated for action improvement. 
The developed DART has several advantages. First, by extracting the task semantics
of LTL formula via Transformer encoder, the agent can more efficiently explore and learn the dexterous manipulation.
Second, instead of simply feeding visual features to the RL module
as in literature, we predict a score map through affordance
learning and plan an optimal manipulation point to guide the interaction with the articulated objects. In addition, we exploit a {diffusion-based policy} that can be learned from scratch, which not only
inherits and promotes the generalization of the diffusion model itself,
but also makes use of the action improvement to enhance the performance.

By exploiting the LTL progression introduced in Sec. \ref{subsec:LTL Progression},
we develop an augmented MDP with an LTL task $\varphi$,
namely task-driven labeled MDP (TL--MDP), as follows.
\begin{defn}
\textbf{\label{Def:TL-MDP}}Given a labeled MDP $\mathcal{M}_{e}=\left(S,T,A,p_{e},\Pi,L,R,\gamma,\mu,\Omega,O\right)$
corresponding to an LTL task $\varphi$, the TL--MDP is constructed
by augmenting $\mathcal{M}_{e}$ to $\mathcal{M_{\varphi}}\triangleq\left(\tilde{S},\tilde{T},A,\tilde{p},\Pi,L,\tilde{R}_{\varphi},\gamma,\mu,\Omega,O\right)$,
where $\tilde{S}=S\times\mathrm{cl}(\varphi)$ with $\mathrm{cl}(\varphi)$ denoting the progression closure of
$\varphi$, i.e., the smallest set containing $\varphi$ that is closed
under progression, $\tilde{T}=\{(s,\varphi)|s\in T,\text{or }\varphi\in\{\mathrm{True},\mathrm{False}\}\}$, $\tilde{p}((s^{'},\varphi^{'})|(s,\varphi),a)=p_{e}(s'|s,a)$
if $\varphi^{'}=\mathrm{prog}(L(s),\varphi)$ and $\tilde{p}((s^{'},\varphi^{'})|(s,\varphi),a)=0$
otherwise, and $\tilde{R}_{\varphi}$ is the reward function associated
with the task $\varphi$ to overcome the non-Markovian reward issue
which is designed as
\begin{equation}
\tilde{R}_{\varphi}(s,\varphi)=\begin{cases}
r_{env}+r_{\varphi}, & \text{if \ensuremath{\mathrm{prog}(L(s),\varphi)=\mathrm{True}}},\\
r_{env}-r_{\varphi}, & \text{if \ensuremath{\mathrm{prog}(L(s),\varphi)=\mathrm{False}}},\\
r_{env}, & \text{otherwise.}
\end{cases}\label{eq:Markovian Reward}
\end{equation}
where $r_{env}$ is the environmental reward, and $r_{\varphi}$ corresponds
to the extra reward when $\sigma$ satisfies the LTL task $\varphi$. 

\end{defn}

The pseudo-code of DART is shown in Alg. \ref{Alg1_DART}, which includes an
exploration phase (lines 4-11) and a training phase (lines 12-16).
In the exploration phase, affordance learning is first used to predict the max-affordance point of manipulation based on the point clouds of the object at the
initial moment (lines 6-7). Meanwhile, LTL representations are encoded using Transformer (line 8). During the interaction,
the {diffusion-based policy} outputs an action $a$ to interact with the
environment guided by vision features and LTL representations (line
9). Then the transition corresponding to the action $a$ is added
to the data set $\hat{B}$ (line 10). In the training phase, the transitions
are not only sampled to update the critic and the {diffusion-based policy},
but also fed to the noise estimator to fit the approximate $\nabla\log p_{t}\left(\cdot\right)$
(lines 13-14). In addition, the state-action pairs $\left(s,a\right)$
in $\hat{B}$ are exploited to optimize the action towards
the reward improvement (line 14). When the manipulation success rate
exceeds the threshold, the contact planner is updated with the contact
positions to predict better interaction points (line 15). Note
that the task representation module, which incorporates the semantics
of the LTL instruction $\varphi$ and encodes the LTL representation,
can be updated indirectly by the gradient backpropagation of RL module.
This process continues until the training budget is exhausted.
Then, DART terminates and returns the
learned policy and network parameters.

\begin{algorithm}
\caption{\label{Alg1_DART}DART}

\scriptsize

\singlespacing

\begin{algorithmic}[1]

\Procedure {Input:} {An LTL instruction $\varphi$ and the MDP
$\mathcal{M}_{e}$ corresponding to some $\varphi$}

{Output: } {\textcolor{red}{{} }Improved action $\hat{a}_{K}$, optimized
Contact Planner $\mathrm{CP}(\cdot)$, noise estimator $\varepsilon_{\xi}\left(\cdot\right)$
and Transformer Encoder in TL-MDP $\mathcal{M}_{\varphi}$}

{Initialization: } {All neural network weights}

\State Load the pretrained weights to the PointNet $\mathrm{PN}(\cdot)$

\While {episode not terminated}

\For {step $t=0,1,...,T$} \Comment{Exploration Phase}

\State $\varphi^{'}\leftarrow\mathrm{prog}(L(s),\varphi)$

\State Extract the vision representation $o_{pn}$ from 3D point
cloud by the PointNet $\mathrm{PN}(\cdot)$

\State Get the max-affordance point observation $o_{\mathrm{MP}}$
from object point clouds $o^{obj}$ by the contact planner $\mathrm{CP}(\cdot)$

\State Augment the observation $o_{0}$ with $\varphi_{\theta}$
encoded by Transformer

\State Determine $\tilde{R}_{\varphi}$ by (\ref{eq:Markovian Reward})
and gather data from $\varphi$ following the {diffusion-based policy} $\pi_{\varphi}$

\State Add the corresponding transition to the data set $\hat{B}$

\EndFor

\For {training step $t=0,1,...,K$} \Comment{Training Phase}

\State Update the noise estimator $\varepsilon_{\xi}\left(\cdot\right)$
in diffusion process by (\ref{eq:estimitor_loss})

\State Improve the action performance by (\ref{eq:action_improvement})
and update the corresponding critic module by (\ref{eq:critic_update})

\State Optimize Contact Planner $\mathrm{CP}(\cdot)$ by (\ref{eq:CP_loss})

\EndFor

\EndWhile

\State Get Optimized neural network weights

\EndProcedure

\end{algorithmic}
\end{algorithm}

In the following, Sec. \ref{subsec:Segmentation} describes the
extraction of point cloud features and the complementation of the
agent's point cloud. Sec. \ref{subsec:LTL Representation} presents
how Transformer extracts the task semantics of the LTL formula and
guides the agent to explore the environment. Sec. \ref{subsec:Affordance Learning}
depicts a series of designs in DART around affordance learning to
improve the success rate of articulated manipulation. Sec. \ref{subsec:Diffusion Policy}
explains how the online {
diffusion-based policy} fits the
reverse process and performs the action improvement.

\subsection{Feature Extraction\label{subsec:Segmentation}}

Previous works typically focus on using images as visual inputs to guide the manipulation of articulated
objects \cite{nair2022r3m}. However, images
often lack necessary spatial information and are vulnerable to perspective changes, resulting in degraded performance. Instead,
the visual features extracted from 3D point clouds contain spatial information that can effectively improve the generalization of the agent's performance
and provide a robust basis for the object manipulation under
different camera viewpoints. Hence, this work
is motivated using 3D point clouds to extract the corresponding
visual features to guide the manipulation.

Specifically, we use PointNet \cite{qi2017pointnet} $\mathrm{PN}(\cdot)$
to extract vision features $o_{pn}$ from 3D point clouds. Since it is difficult to make PointNet output stable using the gradient backpropagation of RL controller, we construct a
dataset to pre-train the PointNet's weights and freeze its
weights when using with RL. In this work, the PointNet is pre-trained  by
the segmentation method and the extracted visual features allow the agent to make basic inferences about the functional
parts of articulated objects. For this purpose, we categorize the
operational scenarios of articulated objects into the following four
groups, i.e., the functional part of the object, the rest of the object,
the dexterous hand, and the robot arm, which are labeled and used for the training of PointNet.

\begin{figure}
\centering{}\includegraphics[scale=0.22]{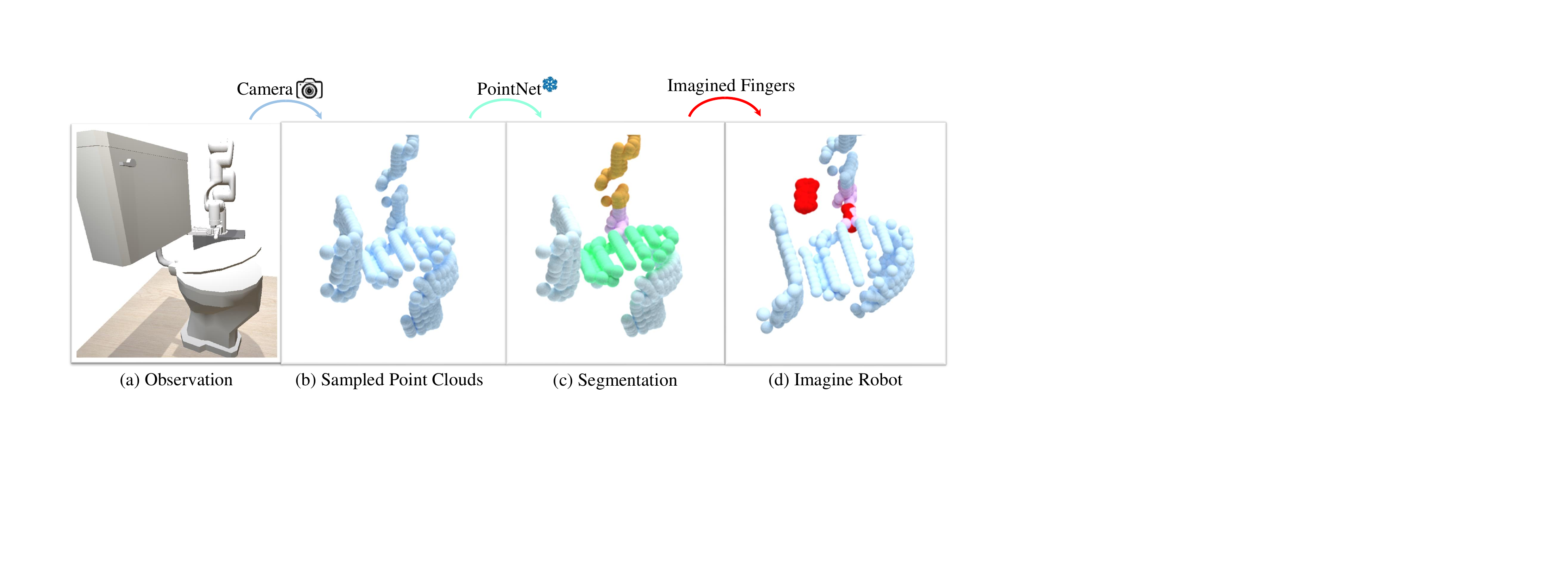}\caption{\textcolor{black}{\label{fig:Outline4PC} A series of processes on
(a) observation containing (b) point cloud sampling, (c) segmentation,
and (d) occluded fingers imagination. The green color represents the functional
part of the toilet in (c), and the red color shows the imagined fingers in
(d). The snowflake indicates the frozen weights.}}
\end{figure}

However, there are two main issues associated with point clouds. One issue is that the occlusion of finger point clouds during interaction may affect the efficiency of learning and the stability
of performance. Another issue is that, when sampling as many points as
possible to avoid loss of important point clouds, the high memory
usage can seriously increase the training time and resource consumption.
To avoid the issues above, we utilize imagined point clouds
to compensate for occluded finger point clouds even if the information
is from a single camera view. Specifically, we first calculate the
pose of each finger via forward kinematics based on the robotic kinematics
model and joint positions. Then, we utilize the potential pose to
sample points from the mesh of each finger link to synthesize
the imagined point cloud of the corresponding finger as shown in
Fig. \ref{fig:Outline4PC}.

\subsection{Task Representation Module based on LTL \label{subsec:LTL Representation}}

In this section, we focus on the use of  LTL representations to understand the task semantics to improve agent\textquoteright{s}
performance. As shown in Fig. \ref{fig:Framework of DART}(a), LTL
progression is first utilized to progress the LTL instruction. {The
operator $\mathrm{prog}$ mentioned in Sec. \ref{subsec:LTL Progression}
takes an LTL formula $\varphi$ and the current label $\sigma_{i}$
as input at each step, and outputs a formula to track which parts
of the original instructions remain to be addressed. For instance,
consider the task $\varphi_{\mathsf{\mathsf{toilet}}}$ in Example
\ref{example1}, which will progress to the subsequent sub-task $\varphi_{\mathsf{\mathsf{toilet}}}^{'}=\lozenge(\mathsf{lid\_grasped}\wedge\lozenge\mathsf{lid\_opened})$
as soon as the robot reaches $\mathsf{toilet\_approached}$.} We
then employ Transformer to encode
LTL representation as part of the input to the RL module, which can indirectly
update the task representation module using the gradient backpropagation
of the reward obtained from the interaction.

Instead of converting LTL specifications to automata or Reward Machine, our previous
work \cite{zhang2023exploiting} also shows the representation encoded
by Transformer \cite{vaswani2017attention}, which provides flexibility
in encoding LTL instructions and facilitates the agent\textquoteright{s} training. Specifically, given an input $X_{\varphi}=(x_{0},x_{1},...)$ generated
by the LTL task $\varphi$ where $x_{t},t=0,1,...,$ represents the
operator or proposition, $X_{\varphi}$ will be preprocessed by the
word embedding $\mathrm{E}$ as $X_{\mathrm{E}}=\left[x_{0}\mathrm{E};x_{1}\mathrm{E};...;x_{N}\mathrm{E}\right]\in\mathbb{R}^{{\normalcolor B\times M\times}D}$
where $B$ is the batch size, $M=N+1$ is the length of input $X_{\varphi}$
and $D$ is the model dimension of the Transformer. And then $X_{\mathrm{E}}$
is added with the frequency-based positional embedding $E_{pos}$
to make use of the order of the sequence. The architecture of Transformer
Encoder for DART is shown in Task Representation Module of Fig. \ref{fig:Framework of DART}(a). Incorporating the LTL representation in TL--MDP is not only appropriate for the forward propagation of the neural network,
but also can be continuously updated as Transformer evolves. 

\subsection{Contact Planner based on Affordance Learning\label{subsec:Affordance Learning}}

To achieve generalizable dexterous manipulation, it is crucial to understand how to effectively manipulate the
functional part. However, to achieve human-level intelligence, previous methods involve either a large number of random contacts in the offline phase
to obtain effective sampling \cite{mo2021where2act}, or human remote control to calibrate
reasonable interaction positions \cite{borja2022affordance}, which greatly increases the learning
cost. {A feasible approach was proposed in \cite{jian2023affordpose}, where the affordance module is trained using human demonstrations. However, collecting such demonstrations for different objects is labor-intensive. Moreover, since the manipulated object in this work is articulated, visualizing the contact posture of all five fingers is challenging, unlike in the case of rigid bodies where finger contact can be easily observed. This difficulty further hinders the efficient learning of the affordance module.}

To address these challenges, we design a contact planner $\mathrm{CP}\left(\cdotp\right)$
that {eliminates the need for offline random acquisition or human demonstrations. Additionally, we develop an effective affordance module that can be optimized through end-to-end online interactions. Instead of fitting contact postures directly, our approach transforms the predicted five-finger contact gesture into a predicted max-affordance point observation (MPO), which guides the palm's interaction.} Specifically, the contact planner
is mainly composed of PointNet++ \cite{qi2017pointnet++} and actionable
affordance module, where PointNet++ mainly extracts visual features
from the segmented point clouds of the object $o^{obj}$, while the
actionable affordance module further maps the visual features into
actionability scores of each point {to indicate the different
potential for successful interaction}, thus converting the point cloud
of an object into a corresponding affordance map. Based on the affordance
map, we select the top $K$ points with the highest actionable
scores and average their positions to obtain the MPO {which represents the interaction
point with the highest success rate}. The agent then uses the MPO as additional information
$o_{\mathrm{MP}}$, which is inputted into the RL controller, to guide
the agent's end-effector $s_{\mathrm{eff}}$ (e.g., the palm position
of the dexterous hand) to interact with the articulated object, thus
achieving effective generalization. In addition, to encourage the
end-effector $s_{\mathrm{eff}}$ to manipulate the functional part
of the articulated object as close as possible to the MPO $o_{\mathrm{MP}}$,
we design the max-affordance point reward (MPR), $r_{\mathrm{MPR}}=1/\left\Vert o_{\mathrm{MP}}-s_{\mathrm{eff}}\right\Vert $,
as a part of $r_{env}$ to indirectly influence
the agent's decision-making.

Through the interaction between the agent and the environment, the
interaction position at which its end-effector completes the task
will further provide the optimized direction for the contact planner's
output. Specifically, during the interaction, we record the contact
points of the end-effector $s_{\mathrm{eff}}$ as $\mathrm{DGT}^{i}(s_{\mathrm{eff}})$
and the corresponding success rates $sr^{i}$ when the articulated object $i$
is manipulated in a fixed number of steps. Using the success rate
$sr^{i}$ as a scale for updating, the following process was designed
to efficiently update and learn the following optimal contact planner
\begin{equation}
\mathrm{CP^{*}}=\underset{\mathrm{CP}}{\mathrm{argmin}}\underset{i}{\sum}sr^{i}\left\Vert \underset{o_{\mathrm{MP}}}{\sum}\mathrm{CP}(o_{\mathrm{MP}}\mid o_{i}^{obj})-\mathrm{DGT}^{i}(s_{\mathrm{eff}})\right\Vert _{2}. \label{eq:CP_loss}
\end{equation}

\subsection{Diffusion Model as Policy Representation for RL\label{subsec:Diffusion Policy}}

{While the visual features extracted in Sec. \ref{subsec:Segmentation}
can provide a basic reasoning ability for the agent's generalization
performance, the quality of the interaction actions is one of the
key factors influencing its manipulation results. Exploiting diffusion
model as a policy representation can improve the generalization performance
of the agent, which has been proven successful in areas such as manipulation.} Exploiting diffusion model as a policy representation
can improve the generalization performance of the agent, which has been proven successful in areas such as manipulation \cite{chi2023diffusion,yang2023policy,shi2024superresolution}.
However, {the generalizability of most diffusion model-based methods in robotics} relies on either solving the problem with offline reinforcement learning (RL) or learning from demonstrations. Such a setting is hard to be applied for dexterous manipulation as related demonstrations are difficult and expensive to collect. Therefore,
in this section, we propose a {diffusion-based policy} without prior
in the context of RL, whose unique action improvement not only addresses 
the above challenges but also ensures generalization.


To simplify the computation mentioned in Sec. \ref{subsec:DM}, we
consider $\mathbf{f}\left(a_{t},t\right)=-\frac{1}{2}g^{2}\left(t\right)a_{t}$
with $g\left(t\right)=\sqrt{2}$ in (\ref{eq: forward in DM}) and
(\ref{eq:reverse in DM}), whose forward noise addition and inverse
denoising processes can be rewritten as $d\bar{a}_{t}=-\bar{a}_{t}dt+\sqrt{2}d\mathbf{w}_{t}$
and 
\begin{equation}
d\tilde{a}{}_{t}=\left[\tilde{a}{}_{t}+2\nabla\log p_{T-t}\left(\tilde{a}{}_{t}\right)\right]dt+\sqrt{2}d\mathbf{\bar{w}}_{t}.\label{eq:reverse process in DP}
\end{equation}
Thus, {as shown in Fig. \ref{fig:Framework of DART}(b)}, we can derive $\bar{a}_{t}$ from a series of forward noise
addition policy distributions $\left\{ \bar{\pi}_{t}\left(\cdot|o\right)\right\} _{t=0:T}$,
where $\bar{\pi}_{0}\left(\cdot|o\right)=\pi\left(\cdot|o\right)$
is the input policy distribution and $\bar{\pi}_{T}\left(\cdot|o\right)$
is standard Gaussian distribution. Similarly, $\tilde{a}{}_{t}$ can
be obtained from a series of inverse denoising policy distributions
$\left\{ \tilde{\pi}_{t}\left(\cdot|o\right)\right\} _{t=0:T}$ in
(\ref{eq:reverse process in DP}), where $\tilde{\pi}_{t}\left(\cdot|o\right)=\bar{\pi}_{T-t}\left(\cdot|o\right)$
if $\tilde{a}_{0}\sim\bar{\pi}_{T}\left(\cdot|o\right)$ and $\nabla\log p_{t}\left(\bar{a}_{t}\right)=\nabla\log\bar{\pi}_{t}\left(\bar{a}_{t}|o\right)$.
For any visual features $o_{pn}$ corresponding to observation $o$,
LTL representation $\varphi_{\theta}$, the max-affordance point
observation $o_{\mathrm{MP}}$ and proprioceptive data $s_{\mathrm{prop}}$, we can design the noise estimator $\varepsilon_{\xi}\left(\bar{a}_{t},o_{pn},\varphi_{\theta},o_{\mathrm{MP}},s_{\mathrm{prop}},t\right)$
to approximate the gradient of the logarithm of the marginal density function $\nabla\log\bar{\pi}_{t}\left(\bar{a}_{t}|o_{pn},\varphi_{\theta},o_{\mathrm{MP}}\right)$
in the dexterous manipulation by

\vspace{-10pt}
\begin{equation*}
\begin{aligned}
L_{\varepsilon}\left(\xi\right) = & \int_{0}^{T}\omega\left(t\right)\underset{\bar{a}_{0}\sim\pi\left(\cdot|o\right),\bar{a}_{t}|\bar{a}_{0}}{E}\left[
\left\Vert \varepsilon_{\xi}\left(\bar{a}_{t},o_{pn},\varphi_{\theta},o_{\mathrm{MP}},s_{\mathrm{prop}},t\right)\right.\right.\\
& \left.\left.-\nabla\log\psi_{t}\left(\bar{a}_{t}|\bar{a}_{0}\right)\right\Vert _{2}^{2}\right]dt
\end{aligned}
\end{equation*}
where $\omega\left(t\right):\left[0,T\right]\rightarrow\mathbb{R}^{+}$ means
a positive weight function, $\psi_{t}\left(\bar{a}_{t}|\bar{a}_{0}\right)$
represents the transition kernel of the forward process (\ref{eq: forward in DM}),
and $\xi$ are parameters in noise estimator $\varepsilon_{\xi}$.
For brevity, we denote the tuple of visual
features $o_{pn}$, LTL representation $\varphi_{\theta}$, the
max-affordance point observation $o_{\mathrm{MP}}$ and proprioceptive data $s_{\mathrm{prop}}$ by $\check{o}$
as $\check{o}=\left(o_{pn},\varphi_{\theta},o_{\mathrm{MP}},s_{\mathrm{prop}}\right)$. When
the agent interacts with the environment, we can obtain the exact
solution from the reverse denoising process (\ref{eq:reverse process in DP})
by performing $K$ iterations to gradually denoise a random noise $\hat{a}_{0}$
into the noise-free action $\hat{a}_{K}$ as
\begin{equation}
\hat{a}_{k+1}=\frac{1}{\sqrt{\alpha_{k}}}\left(\hat{a}_{k}-\frac{1-\alpha_{k}}{\sqrt{1-\bar{\alpha}_{k}}}\varepsilon_{\xi}\left(\hat{a}_{k},\check{o},k\right)\right)+{\sigma}\sqrt{\frac{1-\alpha_{k}}{\alpha_{k}}}z\label{eq:simplied action form}
\end{equation}
 where $z\sim\mathcal{N}\left(0,\mathbf{I}\right)$, ${\sigma\in\mathbb{R}}$,  ${\alpha_{k}\in\mathbb{R}}$,  
and $\bar{\alpha}_{k}=\prod_{i=1}^{k}\alpha_{i}$ are coefficients dependent on the iteration $k$. Then we
can add the corresponding transition to the data set $\hat{B}$. Note
that $T=hK$ where $h\in\mathbb{Z}^{+}$ is the step size to discrete
the interval $\left[0,T\right]$, {and $\sigma=0$ when
deploying and $\sigma=1$ otherwise}.

Compared with the priority-based diffusion policy which only needs
to update the estimator $\varepsilon_{\xi}\left(\cdot\right)$, when
updating {the diffusion-based policy} in DART, we not only need to train
the estimator $\varepsilon_{\xi}\left(\cdot\right)$ to approximate
$\nabla\log p_{t}\left(\cdot\right)$, but also want to sample the
good action $a$ from the Gaussian distribution, so that it can fulfill manipulation tasks as soon as possible. Specifically,
{the update of the policy} in DART can be divided into the
following two steps.

1) Update the estimator $\varepsilon_{\xi}\left(\cdot\right)$: Inspired
by \cite{ho2020denoising}, the loss $L_{\varepsilon}\left(\xi\right)$
can be simplified to 
\begin{equation}
L_{\varepsilon}\left(\xi\right)=\underset{\left(\check{o},a\right)\sim\hat{B}}{E}\left[\left\Vert z-\varepsilon_{\xi}\left(\sqrt{\alpha_{k}}\hat{a}+\sqrt{1-\alpha_{k}}z,\check{o},k\right)\right\Vert _{2}^{2}\right].\label{eq:estimitor_loss}
\end{equation}

2) Action improvement of {diffusion-based policy}: In (\ref{eq:simplied action form}),
since the inverse denoising process to obtain action $\hat{a}$ does not contain
any policy parameters, traditional RL method is difficult to effectively update
the agent's decision. Therefore, we incorporate the action improvement for the following two reasons. On one hand, from
(\ref{eq:simplied action form}), it can be seen that in the $K$ iterations, the action in the previous step plays an
important role in the quality of the action in the next step. On the
other hand, from the RL point of view, the measure of how good an
agent's output action is the reward it receives when interacting with
the environment. To summarize, we propose action improvement: 
\begin{equation}
\hat{a}\leftarrow\hat{a}+\beta\nabla_{a}J(a)=\hat{a}+\beta\underset{\left(\check{o},a\right)\sim\hat{B}}{E}\left[\nabla_{a}Q_{\omega}(\check{o},\hat{a})\right],\label{eq:action_improvement}
\end{equation}
where $\beta$ is the learning rate and $Q_{\omega}\left(o,a\right)$
is the value function parameterized with weights $\omega$ whose
update follows 
\begin{equation}
J_{Q}\left(\omega\right)=\underset{\left(\check{o},a\right)\sim\hat{B}}{E}\left[\left(Q_{\omega}(\check{o},\hat{a})-\left(\tilde{R}_{\varphi}+(\check{o}^{'},\hat{a}^{'})\right)\right)^{2}\right].\label{eq:critic_update}
\end{equation}


















By designing (\ref{eq:action_improvement}), the agent not only overcomes
the limitation of depending on the prior in the previous diffusion
policy, but also takes advantage of multi-modal policy representation
in diffusion process to provide a good basis for generalization. In
addition, although DART does not involve the updating of policy parameters,
the action improvement can ensure convergence, {as demonstrated in \cite{yang2023policy}}.

\subsection{{Advantages Analysis}\label{subsec:Advantages Analysis}}

{The advantages of DART can be summarized as follows.}

{First, compared with traditional visual RL algorithms, which
are usually based on RGB-D images to guide the agent to perform manipulations,
the extracted features affect the agent's spatial reasoning ability.
{Differently, we extract the visual features of the current scene
based on point clouds via PointNet, and provide a generalization basis
for bringing the agent to manipulate on the functional part through
the pre-trained segmentation capability.} We evaluate the effect of
different pretraining methods for PointNet on the generalization and
verify its effectiveness in Sec. \ref{sec:Experiments}. }

{Second, we design a task representation module to extract LTL representations
to guide the agent to improve sampling efficiency. Compared with the
traditional ways of combining automata and RL, encoding LTL instructions
to extract task semantics through Transformer not only builds the
product-MDP on-the-fly to prevent the complexity of the algorithm
from increasing exponentially with the LTL task, but also further
improves the agent's learning efficiency and manipulation ability
through flexible characterization. }

{In addition, we design a contact planner based on affordance learning, which eliminates the need for offline data collection, as required in \cite{mo2021where2act}, making it more compatible with reinforcement learning. Our approach predicts the affordance map using the object's point cloud and plans the optimal contact points $o_{\mathrm{MP}}$ for manipulation through an end-to-end learning framework. Consequently, the agent is guided effectively in manipulating articulated objects by leveraging both the optimal contact point $o_{\mathrm{MP}}$
and the designed reward $r_{\mathrm{MPR}}$, which incentivizes the agent to reach the $o_{\mathrm{MP}}$.}

{Unlike the computationally intensive setup of Meta-RL and the unimodal policy representation of traditional RL algorithms, we adopt a diffusion-based policy as the RL backbone for interacting with the environment. This approach addresses the challenges of using DM for policy representation, such as training from an offline RL perspective or learning from demonstrations. Furthermore, the convergence of the diffusion-based policy provides a theoretical foundation for ensuring both the reliability of action improvement and the effective generalization of DART, as demonstrated in
Sec. \ref{sec:Experiments}.}

{Compared to other generalization methods discussed in Sec. \ref{sec:Intro}, the three modules in DART are tightly integrated, leading to a threefold improvement. Specifically, the LTL representation and MPO guide the diffusion-based policy to generate actions that maximize rewards. Effective interaction with articulated objects enhances the contact planner's ability to predict more accurate MPO, while the resulting actions further refine the Transformer, leading to improved task semantics.}

\section{Experiments\label{sec:Experiments}}

\begin{table*}
\caption{\label{tab:SR4methods}Success Rate of Different Methods on six scenarios
for Both Seen and Unseen Objects.}

\centering{}\resizebox{0.81\textwidth}{!}{
\begin{tabular}{c|cc|cc|cc}
\hline 
Task & \multicolumn{2}{c|}{Toilet} & \multicolumn{2}{c|}{Faucet} & \multicolumn{2}{c}{Laptop}\tabularnewline
\hline 
Split & Seen & Unseen & Seen & Unseen & Seen & Unseen\tabularnewline
\hline 
No Pretrian & 0.710 $\pm$ 0.050 & 0.460 $\pm$ 0.020 & 0.300 $\pm$ 0.220 & 0.280 $\pm$ 0.210 & 0.810 $\pm$ 0.010 & 0.490 $\pm$ 0.090\tabularnewline
DexArt & 0.850 $\pm$ 0.010 & 0.550 $\pm$ 0.010 & 0.790 $\pm$ 0.020 & 0.580 $\pm$ 0.070 & 0.920 $\pm$ 0.020 & 0.600 $\pm$ 0.070\tabularnewline
$\mathrm{DexArt_{aff}}$ & 0.898 $\pm$ 0.051 & 0.597 $\pm$ 0.063 & 0.760 $\pm$ 0.060 & \textbf{0.624 $\pm$ 0.182} & 0.542 $\pm$ 0.048 & 0.533 $\pm$ 0.000\tabularnewline
{QVPO} & {0.306 $\pm$ 0.199} & {0.291 $\pm$ 0.102} & {0.118 $\pm$ 0.092} & {0.095 $\pm$ 0.071} & {0.502 $\pm$ 0.028} & {0.472 $\pm$ 0.051}\tabularnewline
{QSM} & {0.337 $\pm$ 0.092} & {0.208 $\pm$ 0.046} & {0.405 $\pm$ 0.056} & {0.213 $\pm$ 0.091} & {0.396 $\pm$ 0.272} & {0.383 $\pm$ 0.258}\tabularnewline
\hline 
\textbf{DART} & \textbf{0.944 $\pm$ 0.009} & \textbf{0.626 $\pm$ 0.014} & \textbf{0.836 $\pm$ 0.034} & 0.587 $\pm$ 0.022 & \textbf{0.952} $\pm$\textbf{ 0.014} & \textbf{0.872 $\pm$ 0.054}\tabularnewline
\hline 
{Task} & \multicolumn{2}{c|}{{Trashcan}} & \multicolumn{2}{c|}{{Dispenser}} & \multicolumn{2}{c}{{Box}}\tabularnewline
\hline 
{Split} & {Seen} & {Unseen} & {Seen} & {Unseen} & {Seen} & {Unseen}\tabularnewline
\hline 
{No Pretrian} & {0.416 $\pm$ 0.288} & {0.261 $\pm$ 0.184} & {0.730 $\pm$ 0.049} & {0.366 $\pm$ 0.109} & {0.299 $\pm$ 0.050} & {0.294 $\pm$ 0.088}\tabularnewline
{DexArt} & {0.503 $\pm$ 0.122} & {0.411 $\pm$ 0.169} & {0.783 $\pm$ 0.057} & {0.283 $\pm$ 0.061} & {0.780 $\pm$ 0.027} & {0.455 $\pm$ 0.061}\tabularnewline
{$\mathrm{DexArt_{aff}}$} & {0.680 $\pm$ 0.058} & {0.583 $\pm$ 0.040} & {0.776 $\pm$ 0.084} & {0.333 $\pm$ 0.021} & {0.404 $\pm$ 0.035} & {0.333 $\pm$ 0.108}\tabularnewline
{QVPO} & {0.626 $\pm$ 0.092} & {0.511 $\pm$ 0.144} & {0.779 $\pm$ 0.021} & {0.362 $\pm$ 0.040} & {0.213 $\pm$ 0.118} & {0.316 $\pm$ 0.089}\tabularnewline
{QSM} & {0.643 $\pm$ 0.040} & {0.516 $\pm$ 0.027} & {0.820 $\pm$ 0.000} & {0.375 $\pm$ 0.066} & {0.023 $\pm$ 0.006} & {0.061 $\pm$ 0.054}\tabularnewline
\hline 
\textbf{{DART}} & \textbf{{0.733}}{{} $\pm$ }\textbf{{0.040}} & \textbf{{0.633}}{{} $\pm$ }\textbf{{0.026}} & \textbf{{0.880 $\pm$ 0.042}} & \textbf{{0.420 $\pm$ 0.021}} & \textbf{{0.861 $\pm$ 0.006}} & \textbf{{0.710 $\pm$ 0.020}}\tabularnewline
\hline 
\end{tabular}}
\end{table*}

In this section, we examine DART in comparison to previous works across the following aspects: 1) Performance: evaluating whether DART achieves higher success rates on both training and test sets in various scenarios. 2) Affordance: assessing the role of affordance guidance within DART's framework. 3) Representation: analyzing how LTL representation directs the agent in executing dexterous manipulation of articulated objects. 4) Robustness: if DART maintains a high success rate when the camera view deviates from the training configuration.

\subsection{Experimental Setup}

\textit{1) Environments:} DexArt \cite{bao2023dexart}, a benchmark for dexterous manipulation of various articulated objects with high degrees of freedom (DOF), is employed. {Specific details about the reward function can be found on our \href{https://sites.google.com/view/dart0257/} {website}.}

\textit{2) Baselines:} To demonstrate the effectiveness of the DART framework,
we empirically compare it to the following baselines. The first baseline
is DexArt from \cite{bao2023dexart}. Since a possible way to
improve DexArt's generalization performance is to leverage affordance
learning to help the robot understand the task and thus improve
the generalization, we extend DexArt with affordance learning ($\mathrm{DexArt_{aff}}$) in the experiment.  {The third
baseline is QVPO \cite{ding2024diffusionbased}, which
leverages the exploration capabilities and multimodality of diffusion
policies to improve its sample efficiency during online interaction.
The fourth one is QSM \cite{psenka2024learning}, which differentiates through the denoising model and achieves policy convergence via Q-score matching, inherently enabling multimodal exploration in continuous domains.}  In addition, we add the baseline No Pretrain from \cite{bao2023dexart} to represent the importance of
pre-training for PointNet.

{The above baselines and DART are performed on the Ubuntu 20.04 desktop with Intel Core i9 CPU, NVIDIA
3090 GPU and 32GB RAM. More details about baselines and hyper-parameters are shown in \href{https://sites.google.com/view/dart0257/}{website}.}

\subsection{Experimental Results}

\textit{1) Performance Evaluation:} we evaluate the success rates of various methods across three scenarios, presenting the results in Table \ref{tab:SR4methods}. From the table, several observations can be made. (1) The method of extracting visual features via pre-training PointNet achieves a higher success rate compared to non-pre-training. {(2) With the help of affordance learning, DART and
$\mathrm{DexArt_{aff}}$ can get higher success rates than other methods
without it. (3) QSM achieves better generalization performance relative
to QVPO, which may be attributed to the better matching of the object's
pose distribution by Q-score and fewer denoising steps affecting the
ability of QVPO in the generalization task. (4) By understanding the
task semantics from LTL representations and MPO guidance, DART shows
better results relative to other diffusion-based policy representation
methods (QVPO and QSM). (5) DART demonstrates superior performance
in most tasks, showing a higher success rate compared to $\mathrm{DexArt_{aff}}$
and DexArt. This advantage arises because DART employs a {diffusion-based
policy} for decision-making, and its multi-modal policy representation
enhances the agent\textquoteright s performance in complex articulated
object training environments. Notably, DART shows an approximate 45\%
improvement in solving the Unseen set of the Laptop task compared
to DexArt.}.

\begin{table*}
\caption{\label{tab:SR4PC}Success Rate of Different Visual Feature Extraction
Views on four scenarios for Both Seen and Unseen Objects.}

\centering{}\resizebox{0.81\textwidth}{!}{
\begin{tabular}{c|cc|cc|cc}
\hline 
Task & \multicolumn{2}{c|}{Toilet} & \multicolumn{2}{c|}{Faucet} & \multicolumn{2}{c}{Laptop}\tabularnewline
\hline 
Split & Seen & Unseen & Seen & Unseen & Seen & Unseen\tabularnewline
\hline 
No Pretrian & 0.710 $\pm$ 0.050 & 0.460 $\pm$ 0.020 & 0.300 $\pm$ 0.220 & 0.280 $\pm$ 0.210 & 0.810 $\pm$ 0.010 & 0.490 $\pm$ 0.090\tabularnewline
Reconstruction \cite{xie2021generative} & 0.760 $\pm$ 0.030 & 0.520 $\pm$ 0.030 & 0.350 $\pm$ 0.020 & 0.210 $\pm$ 0.030 & 0.850 $\pm$ 0.040 & 0.540 $\pm$ 0.080\tabularnewline
SimSiam \cite{chen2021exploring} & 0.820 $\pm$ 0.020 & 0.500 $\pm$ 0.060 & 0.600 $\pm$ 0.150 & 0.450 $\pm$ 0.120 & 0.840 $\pm$ 0.040 & 0.490 $\pm$ 0.130\tabularnewline
\textbf{Segmentation} & \textbf{0.944 $\pm$ 0.009} & \textbf{0.626 $\pm$ 0.014} & \textbf{0.836 $\pm$ 0.034} & \textbf{0.587 $\pm$ 0.022} & \textbf{0.952} $\pm$\textbf{ 0.014} & \textbf{0.872 $\pm$ 0.054}\tabularnewline
\hline 
{Task} & \multicolumn{2}{c|}{{Trashcan}} & \multicolumn{2}{c|}{{Dispenser}} & \multicolumn{2}{c}{{Box}}\tabularnewline
\hline 
{Split} & {Seen} & {Unseen} & {Seen} & {Unseen} & {Seen} & {Unseen}\tabularnewline
\hline 
{No Pretrian} & {0.416 $\pm$ 0.288} & {0.261 $\pm$ 0.184} & {0.730 $\pm$ 0.049} & {0.445 $\pm$ 0.006} & {0.299 $\pm$ 0.050} & {0.294 $\pm$ 0.088}\tabularnewline
{Reconstruction \cite{xie2021generative}} & {0.323 $\pm$ 0.232} & {0.182 $\pm$ 0.124} & {0.736 $\pm$ 0.028} & {0.408 $\pm$ 0.021} & {0.642 $\pm$ 0.084} & {0.655 $\pm$ 0.126}\tabularnewline
{SimSiam \cite{chen2021exploring}} & {0.470 $\pm$ 0.243} & {0.372 $\pm$ 0.177} & {0.739 $\pm$ 0.008} & {0.350 $\pm$ 0.040} & {0.537 $\pm$ 0.070} & {0.638 $\pm$ 0.061}\tabularnewline
\textbf{{Segmentation}} & \textbf{{0.733}}{{} $\pm$ }\textbf{{0.040}} & \textbf{{0.633}}{{} $\pm$ }\textbf{{0.026}} & \textbf{{0.880 $\pm$ 0.042}} & \textbf{{0.420 $\pm$ 0.021}} & \textbf{{0.861 $\pm$ 0.006}} & \textbf{{0.710 $\pm$ 0.020}}\tabularnewline
\hline 
\end{tabular}}
\end{table*}

\begin{figure}
\centering{}\includegraphics[scale=0.215]{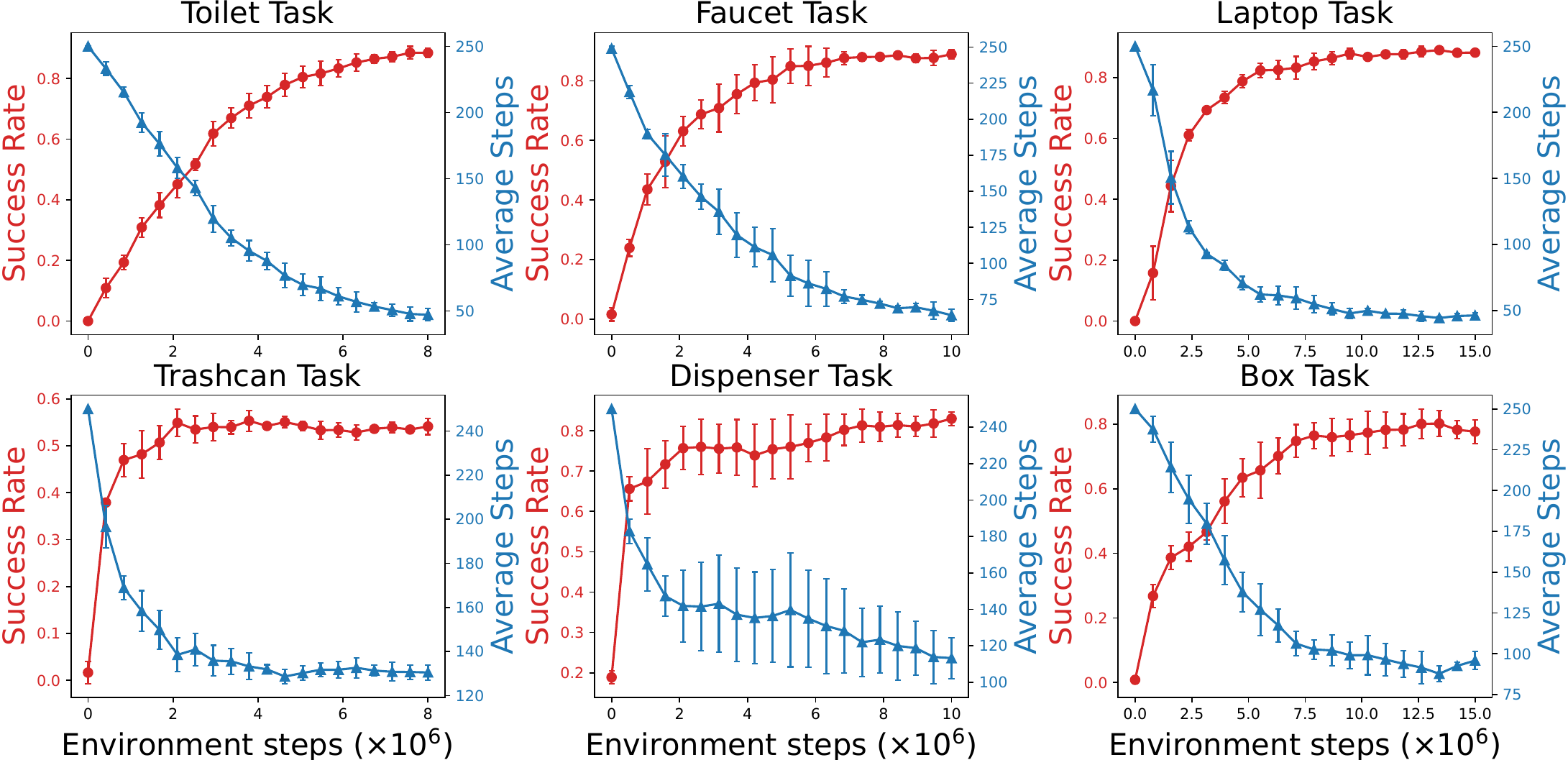}\caption{\label{fig:succ_rate and average_steps}Success rate (Red) and average
success steps (blue) of DART for different tasks during training.}
\end{figure}

\textit{2) Success Rate and Average Success Steps:} We also present the success rate (red) and average success steps (blue) of DART for various tasks during training in Fig. \ref{fig:succ_rate and average_steps}. It can be observed that DART tends to converge at approximately half the total number of training steps, achieving a higher success rate and requiring fewer steps to complete the tasks.

\textit{3) Pre-training of {PointNet}:} The segmentation
pre-training method described in Sec. \ref{subsec:Segmentation} is used to extract
point cloud features and provide the agent with fundamental reasoning about the functional parts of articulated objects. To evaluate the effect of different pre-training methods on PointNet, we also consider Reconstruction \cite{xie2021generative} and SimSiam \cite{chen2021exploring}, presenting their corresponding success rates in Table \ref{tab:SR4PC}. From the table, we observe that 1) the pre-trained PointNet method using segmentation exhibits the highest success rate across all three scenarios. 2) Compared to other tasks, the segmentation method achieves a higher success rate in Laptop task.

\subsection{{Ablation Study}}

\begin{table*}
\caption{\label{tab:ablation4dart_six}Success rates corresponding to ablating
different DART components on six scenarios.}

\centering{}\resizebox{0.81\textwidth}{!}{
\begin{tabular}{c|cc|cc|cc}
\hline 
Task & \multicolumn{2}{c|}{Toilet} & \multicolumn{2}{c|}{Faucet} & \multicolumn{2}{c}{Laptop}\tabularnewline
\hline 
Split & Seen & Unseen & Seen & Unseen & Seen & Unseen\tabularnewline
\hline 
$\mathrm{DART_{w/o}^{aff+LTL}}$ & 0.810 $\pm$ 0.084 & 0.461$\pm$ 0.067 & 0.628 $\pm$ 0.118 & 0.330 $\pm$ 0.114 & 0.840 $\pm$ 0.068 & 0.731 $\pm$ 0.115\tabularnewline
$\mathrm{DART_{w/o}^{aff}}$ & 0.863 $\pm$ 0.033 & 0.537 $\pm$ 0.018 & 0.716 $\pm$ 0.137 & 0.456 $\pm$ 0.099 & 0.922 $\pm$ 0.038 & 0.826 $\pm$ 0.037\tabularnewline
{$\mathrm{DART_{w/o}^{LTL}}$} & {0.715 $\pm$ 0.035} & {0.433 $\pm$ 0.022} & {0.675 $\pm$ 0.092} & {0.388 $\pm$ 0.126} & {0.948 $\pm$ 0.022} & {0.863 $\pm$ 0.029}\tabularnewline
\hline 
\textbf{DART} & \textbf{0.944 $\pm$ 0.009} & \textbf{0.626 $\pm$ 0.014} & \textbf{0.836 $\pm$ 0.034} & \textbf{0.587 $\pm$ 0.022} & \textbf{0.952} $\pm$\textbf{ 0.014} & \textbf{0.872 $\pm$ 0.054}\tabularnewline
\hline 
{Task} & \multicolumn{2}{c|}{{Trashcan}} & \multicolumn{2}{c|}{{Dispenser}} & \multicolumn{2}{c}{{Box}}\tabularnewline
\hline 
{Split} & {Seen} & {Unseen} & {Seen} & {Unseen} & {Seen} & {Unseen}\tabularnewline
\hline 
{$\mathrm{DART_{w/o}^{aff+LTL}}$} & {0.683 $\pm$ 0.073} & {0.588 $\pm$ 0.028} & {0.846 $\pm$ 0.067} & {0.308 $\pm$ 0.021} & {0.466 $\pm$ 0.013} & {0.616 $\pm$ 0.013}\tabularnewline
{$\mathrm{DART_{w/o}^{aff}}$} & {0.716 $\pm$ 0.032} & {0.621 $\pm$ 0.102} & {0.840 $\pm$ 0.035} & {0.404 $\pm$ 0.005} & {0.571 $\pm$ 0.046} & {0.599 $\pm$ 0.062}\tabularnewline
{$\mathrm{DART_{w/o}^{LTL}}$} & {0.726 $\pm$ 0.018} & {0.625 $\pm$ 0.042} & {0.656 $\pm$ 0.033} & {0.125 $\pm$ 0.020} & {0.694 $\pm$ 0.088} & {0.511 $\pm$ 0.083}\tabularnewline
\hline 
\textbf{{DART}} & \textbf{{0.733}}{{} $\pm$ }\textbf{{0.040}} & \textbf{{0.633}}{{} $\pm$ }\textbf{{0.026}} & \textbf{{0.880 $\pm$ 0.042}} & \textbf{{0.420 $\pm$ 0.021}} & \textbf{{0.861 $\pm$ 0.006}} & \textbf{{0.710 $\pm$ 0.020}}\tabularnewline
\hline 
\end{tabular}}
\end{table*}
{\textit{1) The impact of different components:} We design
the following ablation study to evaluate the influence of different
components in DART. (1) $\mathrm{DART_{w/o}^{aff+LTL}}$ is DART
without considering affordance learning and LTL representations. (2)
$\mathrm{DART_{w/o}^{aff}}$is DART without considering affordance
learning to show the importance of MPO $o_{\mathrm{MP}}$ and MPR
$r_{\mathrm{MPR}}$. (3) $\mathrm{DART_{w/o}^{LTL}}$ is DART without
considering LTL representations to show the importance of task semantics
understanding.}

{As shown in Table \ref{tab:ablation4dart_six}, several observations can be made. (a)
Without the help of the affordance learning and LTL representations,
$\mathrm{DART_{w/o}^{aff+LTL}}$ get the worse {performance} on all scenarios.
(b) By representing the LTL specifications, DART achieves better generalization
results compared to $\mathrm{DART_{w/o}^{LTL}}$. In the task representation
module, the design of a Transformer Encoder to encode LTL instructions
not only facilitates the backpropagation of neural networks but also
leverages the extracted task semantics. (c) By predicting MPO from
the contact planner to guide the agent, DART shows better generalization
performance compared to $\mathrm{DART_{w/o}^{aff}}$.}

\begin{figure}
\centering{}\includegraphics[scale=0.21]{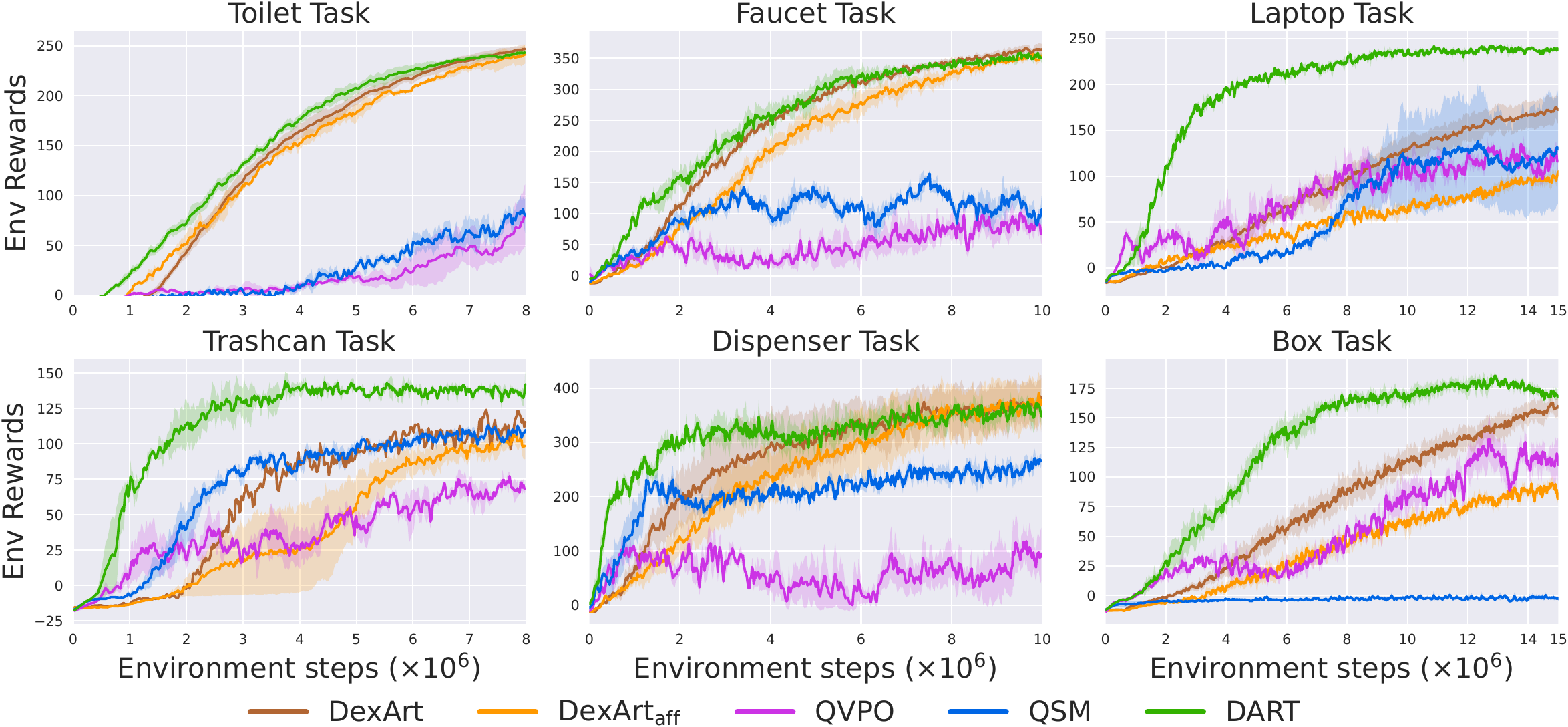}\caption{\label{fig:ablation4DART}Different Methods learning performing in seen objects of scenarios.}
\end{figure}

\begin{figure}
\centering{}\includegraphics[scale=0.215]{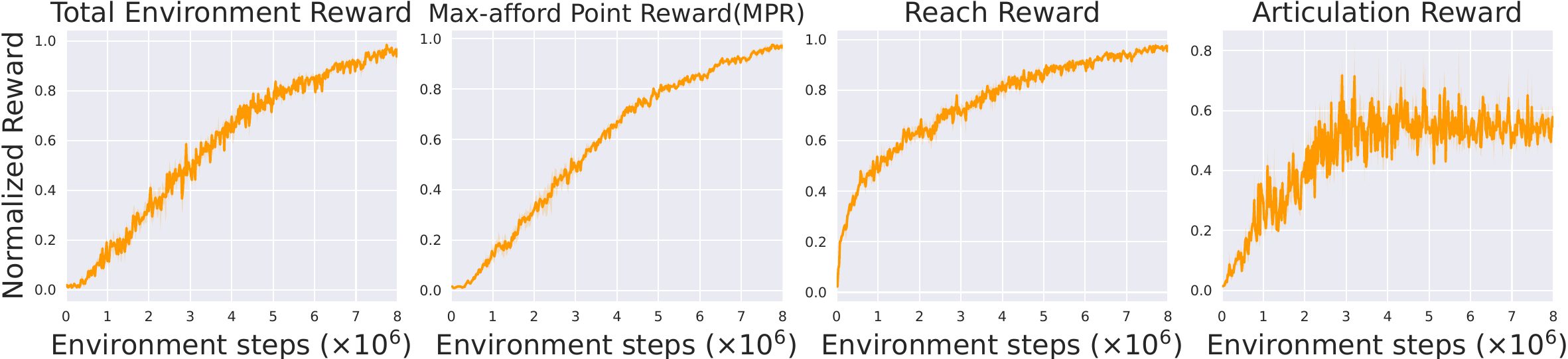}\caption{\label{fig:reward_detailed}Normolized reward components for DART
in the Toilet task .}
\end{figure}

{2)} {\textit{Key reward components visualization:}
As shown in Fig. \ref{fig:reward_detailed}, we present the training
process of the key reward components of DART in the Toilet task. It
can be observed that (1) the increasing trend of MPR is similar to
the curve of success rate, which means that the increasing MPR contributes
to the success rate of the manipulation; (2) the articulation reward
stabilizes around three million steps, which represents that the agent
can succeed in most of the Toilet objects since that; and (3) the
reach reward increases with the training process, which shows that
the agent uses fewer steps to quickly approach the articulated
object within a certain number of steps.}

\begin{table}
\caption{\label{tab:SR_task_representation}Success Rate of Different Task
Representation Methods.}

\centering{}\resizebox{0.45\textwidth}{!}{
\begin{tabular}{c|cc|c}
\hline 
{Task} & \multicolumn{2}{c|}{{Faucet}} & \multicolumn{1}{c}{{Dispenser}}\tabularnewline
\hline 
{Split} & {Seen} & {Unseen} & \multicolumn{1}{c}{{Unseen (Transfer)}}\tabularnewline
\hline 
{DFA} & {0.496 $\pm$ 0.038} & {0.347 $\pm$ 0.088} & {0.266 $\pm$ 0.056}\tabularnewline
{LSTM} & {0.621 $\pm$ 0.100} & {0.409 $\pm$ 0.099} & {0.337 $\pm$ 0.056}\tabularnewline
{GRU} & {0.550 $\pm$ 0.193} & {0.442 $\pm$ 0.141} & {0.266 $\pm$ 0.092}\tabularnewline
{R-GCN} & {0.433 $\pm$ 0.118} & {0.290 $\pm$ 0.059} & {0.257 $\pm$ 0.065}\tabularnewline
\hline 
{Transformer (ours)} & \textbf{{0.933 $\pm$ 0.010}} & \textbf{{0.639 $\pm$ 0.026}} & \textbf{{0.400 $\pm$ 0.044}}\tabularnewline
\hline 
\end{tabular}}
\end{table}

\begin{figure}
\centering{}\includegraphics[scale=0.13]{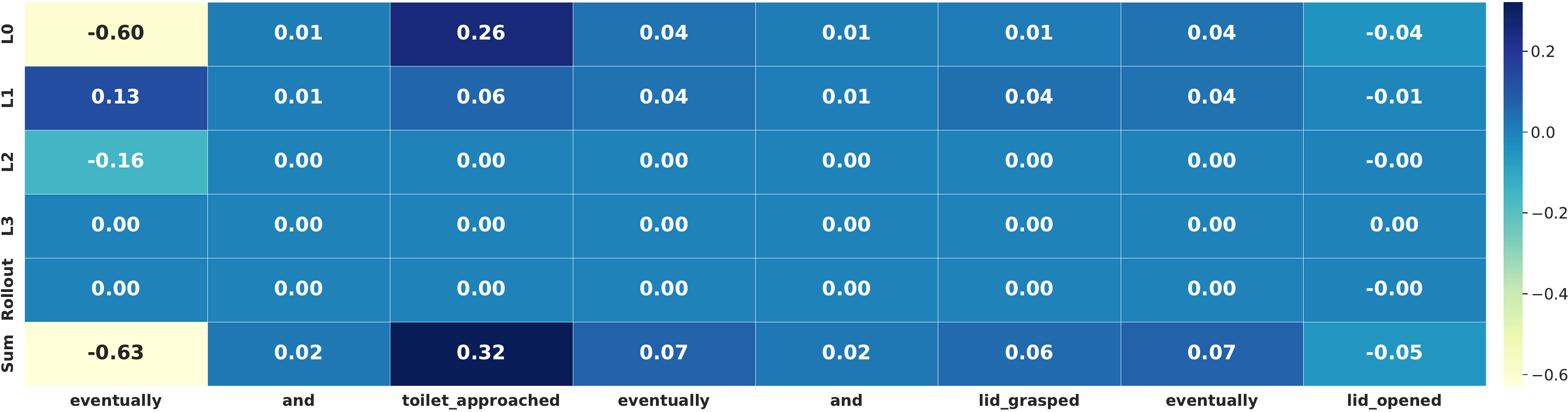}\caption{\label{fig:AttCAT}The heatmap of the task $\varphi_{\mathsf{\mathsf{toilet}}}$
by normalizing impact scores from different Transformer layers.}
\end{figure}

\begin{table}
\caption{\label{tab:time}Total training time (hours).}

\centering{}\resizebox{0.45\textwidth}{!}{
\begin{tabular}{c|c|c|c|c|c|c}
\hline 
Task & Toilet & Faucet & Laptop & Trashcan & Dispenser & Box\tabularnewline
\hline 
DexArt & \textbf{10.763} & 1\textbf{5.338} & 24.509 & 12.277 & 14.854 & 23.148\tabularnewline
$\mathrm{DexArt_{aff}}$ & 11.055 & 15.605 & 25.252 & 12.484 & 15.605 & 23.946\tabularnewline
\hline 
DART & 11.883 & 15.432 & \textbf{23.020} & \textbf{11.574} & \textbf{14.172} & \textbf{22.281}\tabularnewline
\hline 
\end{tabular}}
\end{table}

{\textit{3) The impact of task representation:}
To demonstrate the effect of task representation on DART, we compare
the performance of the following five task representation methods
in DART in the Faucet task. The first method is DFA  \cite{Lacerda2014}, which
converts the task specification into the corresponding automata at
the task level to fire critical task states to guide the agent. The
second and third methods are LSTM and GRU, respectively,
which effectively capture task information over time as text-based methods. The fourth is R-GCN \cite{vaezipoor2021ltl2action}, which
utilizes explicit relational and structural information in the graph
helping the agent to show better generalization ability. And the last
one is our method.}

{As shown in Table \ref{tab:SR_task_representation},
it can be observed that 1) the method of encoding LTL instructions
through Transformer achieves the highest success rate for the training
set, the unseen set, and for the different category; 2) when using
R-GCN as the task representation, the agent achieves the worst generalization
performance, and it may be due to the over-smoothing by designing
multiple message-passing layers; 3) the agent shows the second best
task performance in all scenarios by exploiting the text-based method
as the task representation, which means that text-based task representation
can effectively capture long-term dependencies and understand the
context within tasks.}

{\textit{4) The visualization of task representation:} To further
illustrate the agent\textquoteright s understanding of the LTL task,
Fig. \ref{fig:AttCAT} illustrates the heatmap of the task $\varphi_{\mathsf{\mathsf{toilet}}}$ by exploiting AttCAT \cite{qiang2022attcat}.
Upon the convergence of Transformer, higher impact scores from all
layers focus on the token $\mathsf{toilet\text{\_}apporached}$ (+0.32)
as shown in Fig. \ref{fig:AttCAT}, which suggests that the agent
is more likely to move directly to the position corresponding to the
$\mathsf{toilet\text{\_}apporached}$ proposition.}

{\textit{5) The comparison of time consumption:} To further
clarify the training efficiency of DART at the time level, we evaluate
the average time consumed by DexArt, $\mathrm{DexArt_{aff}}$ and
DART on the above six tasks. As shown in Fig. \ref{fig:ablation4DART}
and Table. \ref{tab:time}, we observe that 1) when the tasks are
simple (e.g., the Toilet and Faucet tasks), the three methods exhibit
similar reward curves, and DexArt consumes the least average time
due to its simplicity. 2) When the tasks are more challenging, DART
not only demonstrates efficient learning performance, but also uses
the shortest training time.}


\subsection{Transfer Reasoning of Unseen Categories}

In \cite{bao2023dexart}, a dexterous manipulation task benchmark
is proposed to evaluate the ability of generalizing across articulated
objects within the same category. We believe that an algorithm capable of effectively reasoning about the functional parts of articulated objects should be able to generalize well to objects within the same category and also exhibit the ability to generalize to objects across different categories. 
Therefore, we extend the categories of articulated objects beyond those in \cite{bao2023dexart}, as detailed on our \href{https://sites.google.com/view/dart0257/} {website},
to evaluate the generalization performance across different categories.

Taking the Trashcan task as an example, we train the agent on the
set of seen objects in the Toilet task, and further test its success
rate for unseen object types in Trashcan task, denoted as Trashcan (Toilet). The same applies to Faucet (Dispenser) and Box (Laptop).
We evaluate the success rate of different methods when transfering to
different category-level objects and show the results in Table \ref{tab:SR4Transfer}.

\begin{table}
\caption{\label{tab:SR4Transfer}Success Rate Evaluation of Different Methods
When Transfering to Different Category-level Objects.}

\centering{}\resizebox{0.45\textwidth}{!}{
\begin{tabular}{c|c|c|c}
\hline 
{Task (}\textbf{{Full}}{)} & {Trashcan (Toilet)} & {Dispenser (Faucet)} & {Box (Laptop)}\tabularnewline
\hline 
{DexArt} & {0.364 $\pm$ 0.122} & {0.246 $\pm$ 0.014} & {0.178 $\pm$ 0.003}\tabularnewline
{$\mathrm{DexArt_{aff}}$} & {0.447 $\pm$ 0.060} & {0.438 $\pm$ 0.031} & {0.243 $\pm$ 0.243}\tabularnewline
{QVPO} & {0.210 $\pm$ 0.129} & {0.353 $\pm$ 0.103} & {0.256 $\pm$ 0.031}\tabularnewline
{QSM} & {0.224 $\pm$ 0.088} & {0.434 $\pm$ 0.081} & {0.222 $\pm$ 0.100}\tabularnewline
\hline 
\textbf{{DART}} & \textbf{{0.562 $\pm$ 0.051}} & \textbf{{0.518 $\pm$ 0.025}} & \textbf{{0.435 $\pm$ 0.072}}\tabularnewline
\hline 
{Task (}\textbf{{Heterogeneous}}{)} & {Trashcan (Toilet)} & {Dispenser (Faucet)} & {Box (Laptop)}\tabularnewline
\hline 
{DexArt} & {0.393 $\pm$ 0.031} & {0.241 $\pm$ 0.025} & {0.144 $\pm$ 0.008}\tabularnewline
{$\mathrm{DexArt_{aff}}$} & {0.388 $\pm$ 0.064} & {0.233 $\pm$ 0.062} & {0.199 $\pm$ 0.084}\tabularnewline
{QVPO} & {0.127 $\pm$ 0.034} & {0.138 $\pm$ 0.056} & {0.044 $\pm$ 0.031}\tabularnewline
{QSM} & {0.038 $\pm$ 0.020} & {0.349 $\pm$ 0.056} & {0.077 $\pm$ 0.039}\tabularnewline
\hline 
\textbf{{DART}} & \textbf{{0.472 $\pm$ 0.103}} & \textbf{{0.400 $\pm$ 0.044}} & \textbf{{0.361 $\pm$ 0.088}}\tabularnewline
\hline 
{Task (}\textbf{{Complex}}{)} & {Trashcan (Toilet)} & {Dispenser (Faucet)} & {Box (Laptop)}\tabularnewline
\hline 
{DexArt} & {0.354 $\pm$ 0.032} & {0.234 $\pm$ 0.005} & {0.145 $\pm$ 0.027}\tabularnewline
{$\mathrm{DexArt_{aff}}$} & {0.320 $\pm$ 0.054} & {0.321 $\pm$ 0.039} & {0.153 $\pm$ 0.022}\tabularnewline
{QVPO} & {0.085 $\pm$ 0.052} & {0.271 $\pm$ 0.037} & {0.194 $\pm$ 0.036}\tabularnewline
{QSM} & {0.078 $\pm$ 0.051} & {0.328 $\pm$ 0.038} & {0.148 $\pm$ 0.050}\tabularnewline
\hline 
\textbf{{DART}} & \textbf{{0.406 $\pm$ 0.026}} & \textbf{{0.395 $\pm$ 0.029}} & \textbf{{0.281 $\pm$ 0.036}}\tabularnewline
\hline 
\end{tabular}}
\end{table}

\begin{figure}
\centering{}\includegraphics[scale=0.18]{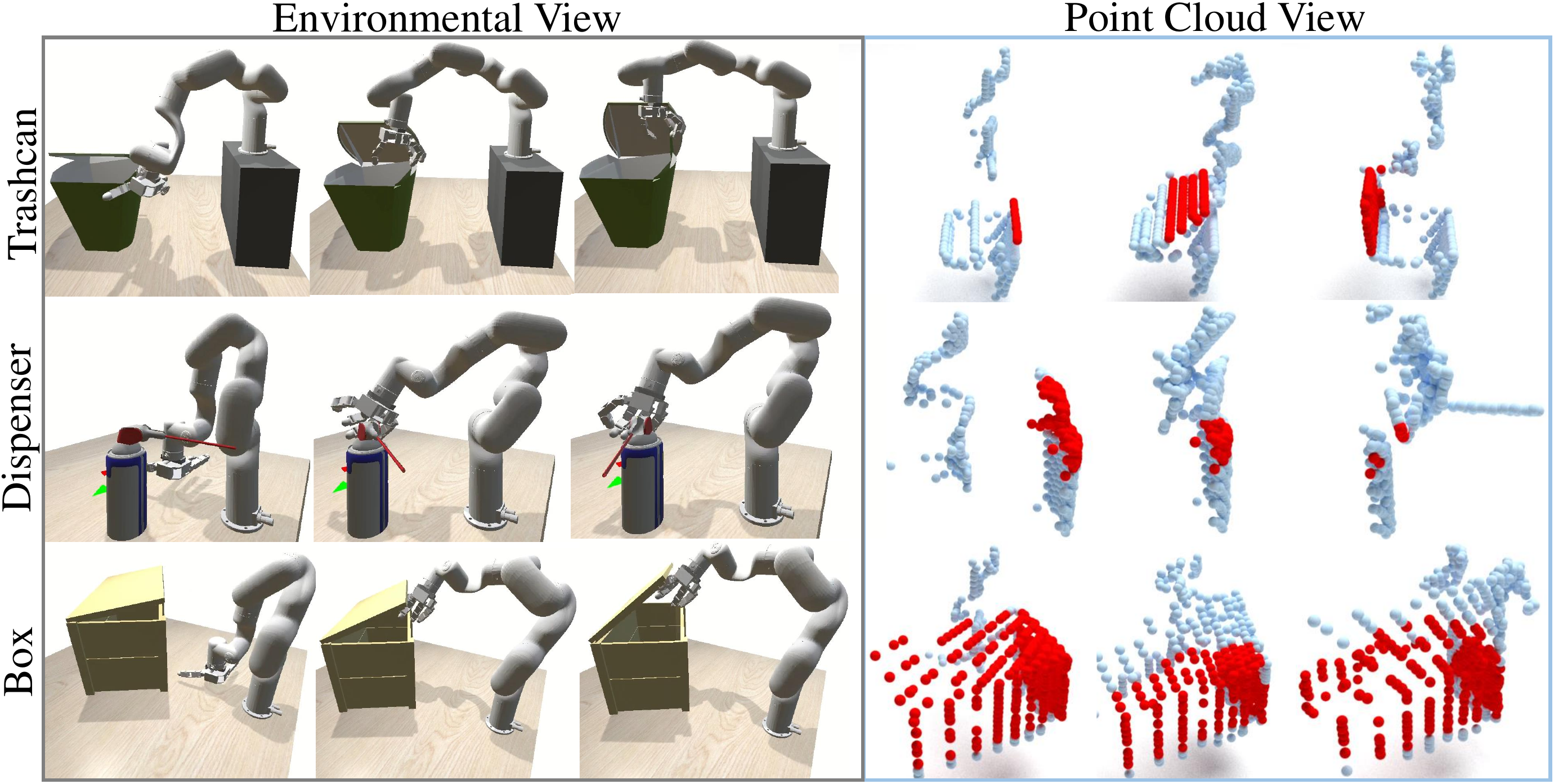}\caption{\label{fig:snapshot4different category}Snapshots from the environment
views and the point cloud views. The completed point cloud views are shown in our \href{https://sites.google.com/view/dart0257/} {website}.}
\end{figure}

From Table \ref{tab:SR4Transfer}, it can be observed that {1) compared
with Table \ref{tab:SR4methods}, DART still shows good generalization
even different category-level objects; 2) Although DART shows lower
success rates in Heterogeneous and Complex sets, compared with other
algorithms, it still has a clear lead advantage. 3) Although QSM and
QVPO are also diffusion-based policy representation methods, they
have shown better results only on the Dispenser and Trashcan tasks,
which may be due to the capacity limitation of Q-Score matching and
the small number of diffusion steps; 4) By predicting MPO through
the contact planner, affordance learning-based algorithms ($\mathrm{DexArt_{aff}}$
and DART) generally achieve a good success rate; 5) DART shows the
highest success rate in all scenarios, and this phenomenon confirms
that the generalization of {diffusion-based policy} can be further improved
under the guidance of task semantics and MPO.} The snapshots
for one of different category-level generalizations are shown in  Fig. \ref{fig:snapshot4different category}.

\subsection{Robustness to Viewpoint Change}

\begin{figure}
\centering{}\includegraphics[scale=0.46]{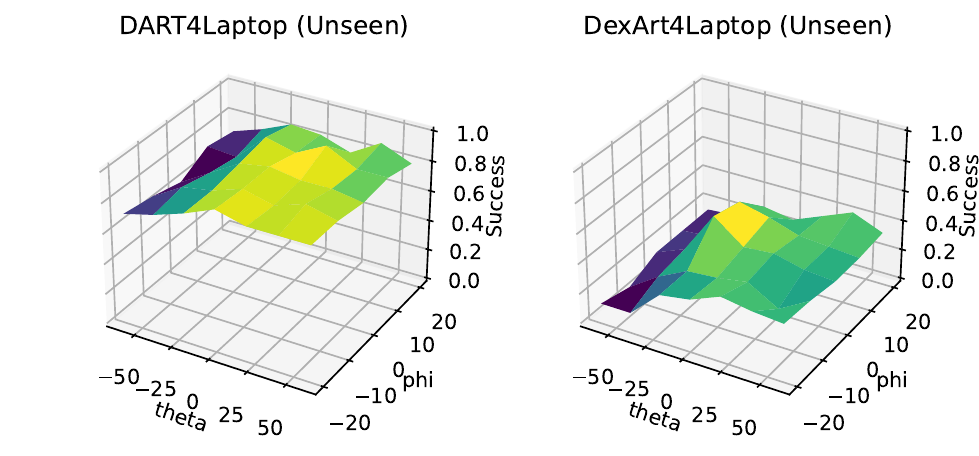}\caption{\label{fig:robust4laptop}We evaluate Success Rate on Laptop task
under Different Viewpoints. The $x$-axis is the polar angle
$\phi$ (relative to the training viewpoint) and the y-axis is the
azimuthal angle $\theta$ on the semi-sphere centered at the object.
The $z$-axis represents the success rate. }
\end{figure}

We further test the robustness of DART when changing the camera viewpoints
in Laptop task following the approach of \cite{bao2023dexart}. The success rate for the unseen set corresponding to a total of 35 camera views is shown in Fig. \ref{fig:robust4laptop}.

As shown in Fig. \ref{fig:robust4laptop}, DexArt, which already exhibits a low success rate on the unseen set, shows a significant decline in success rate as the azimuthal angle changes, reaching only about 10\%. In contrast, DART demonstrates much more robust performance, maintaining a success rate of over 70\% across all 35 views, and achieving over 85\% success for azimuths ranging from -$20^{\text{\textopenbullet}}$ to $60^{\text{\textopenbullet}}$. Additionally, while the training view is the optimal manipulation view for DexArt, DART manages to achieve higher success rates than the training view even as the azimuthal and polar angles vary. 


\subsection{Real--World Experimental Results}

\begin{figure}
\centering{}\includegraphics[scale=0.13]{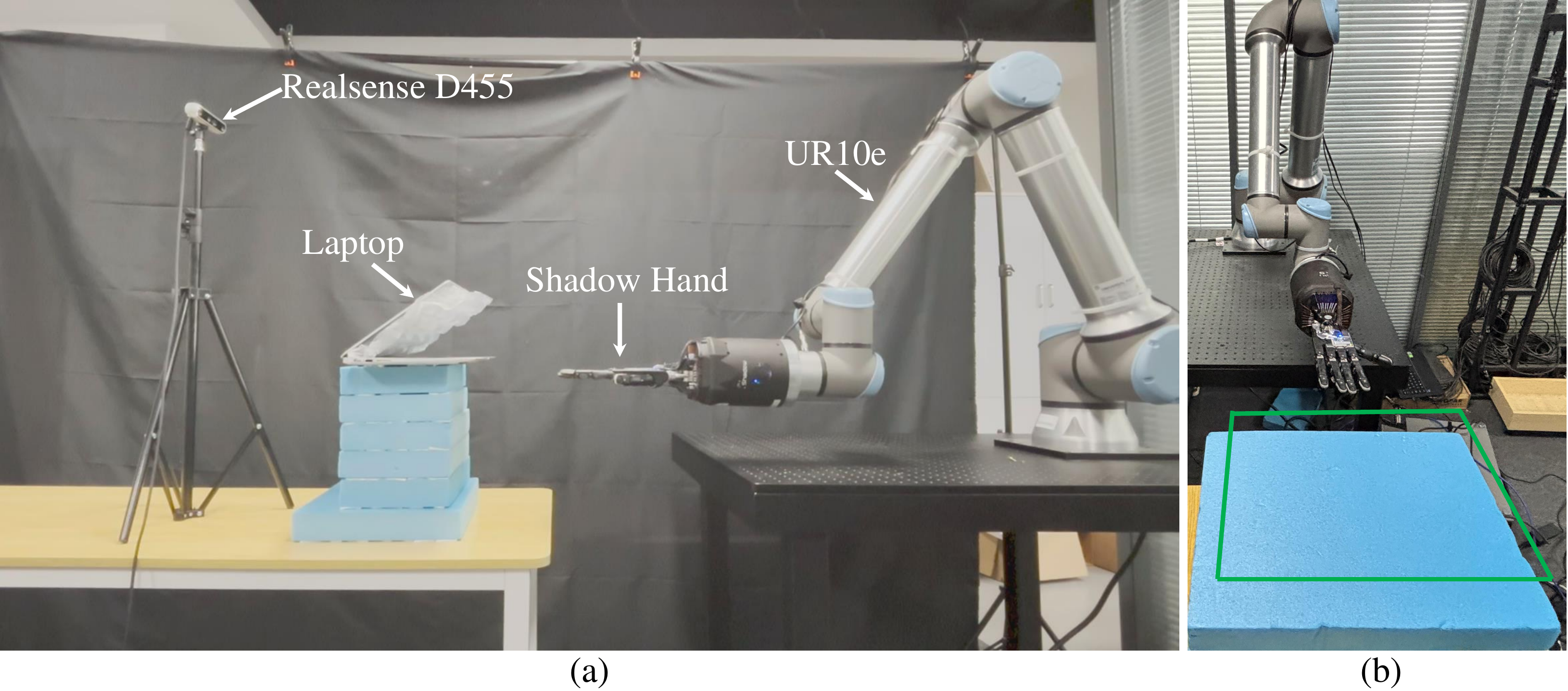}\caption{\label{fig:real-world experimental setting}(a) The real-world experimental
platform constructed for Task Laptop. (b) The green frame means randomized object poses.}
\end{figure}

\begin{figure}
\centering{}\includegraphics[scale=0.16]{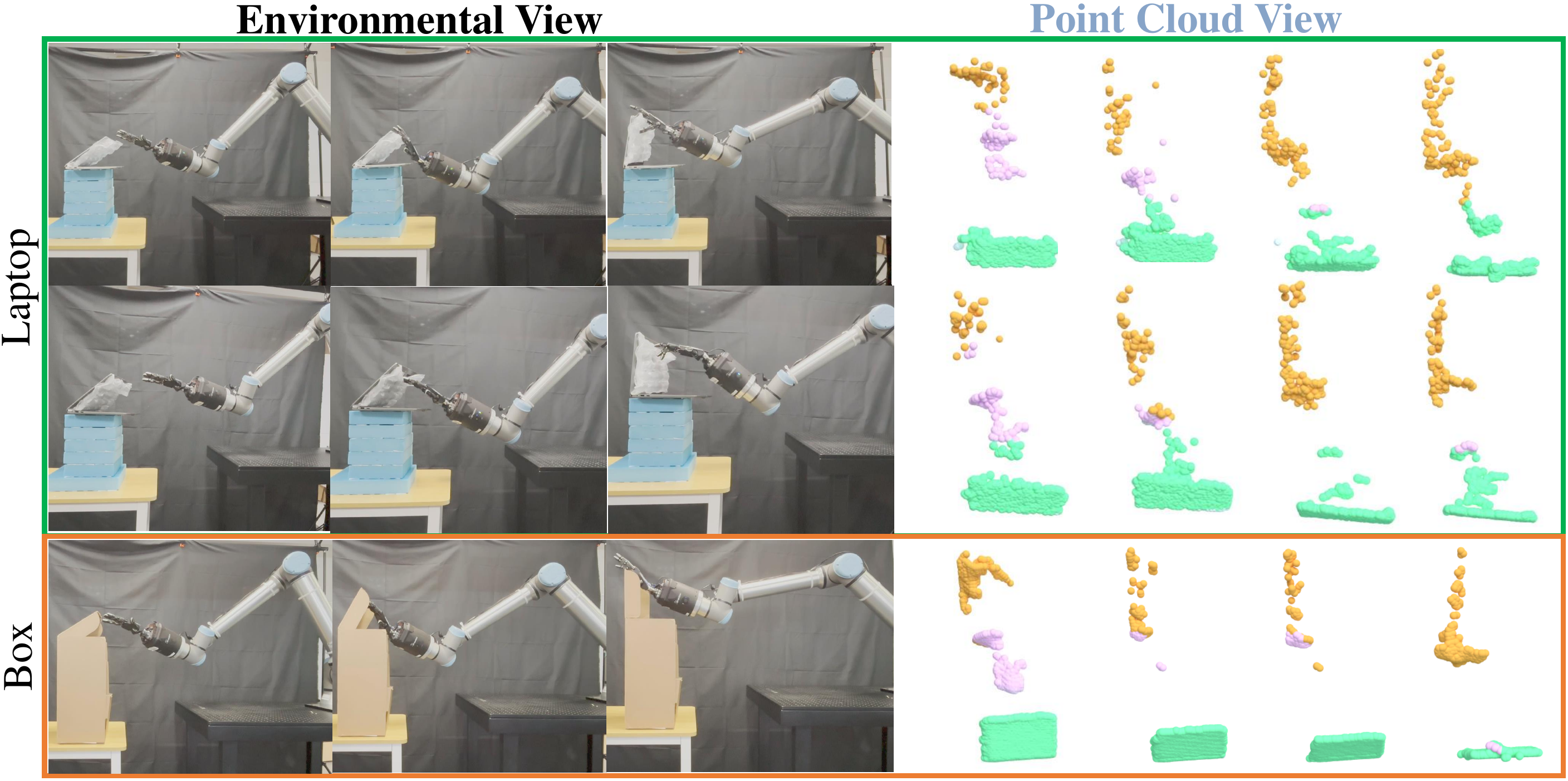}\caption{\label{fig:snapshot4real-world}Snapshots and the corresponding point
cloud views for Task Laptop.}
\end{figure}

To further validate the generalization of DART, physical experiments
were conducted, as illustrated in Fig. \ref{fig:real-world experimental setting}. {We utilize
Realsense D455 to capture point clouds, and exploit the UR10e arm
and Shadow Hand to execute the action output by different methods.
The corresponding results are all performed on the Ubuntu 20.04 laptop
with Intel Core i7-12800HX CPU, NVIDIA 3080TI GPU and 32GB RAM. We train
the performance of the UR10e arm and Shadow Hand in the Laptop task
in simulation, and deploy the corresponding weights directly in the
real world to evaluate the performance that the weights have. It is
worth noting that we do not further adapt or optimize the corresponding
weights during deployment, and the observed noise in the environment
will challenge the performance of the algorithm, thus further demonstrating
the robustness and generalizability of DART.} We first assess the
performance of DART with two laptops and a box. The snapshots and
corresponding point cloud views are shown in Fig. \ref{fig:snapshot4real-world}.
And{{} we then evaluate the success of different methods
by randomly initializing the position and orientation on the Laptop
task as shown in Table \ref{tab:real_evaluation}.
}A full video of the experiment is available on our \href{https://sites.google.com/view/dart0257/}{website}.

{\textit{1) Manipulation effectiveness:} It can be observed
that (1) DexArt and $\mathrm{DexArt_{aff}}$ have a low success rate
because, under the same seed, although they have a certain success
rate in the simulation, they have to be terminated artificially early
due to the dangerous nature of the corresponding actions in the real
world. (2) QVPO has poor performance in real-world scenarios because
it is difficult to succeed in simulation evaluation. (3) QSM has a
second better performance in the real-world scenarios by exploiting
through Q-score matching and differentiating through the denoising
model. (4) DART not only has the highest success rate in both scenarios,
but also the actions are more anthropomorphic than other methods.}

\begin{table}
\caption{\label{tab:real_evaluation}Success rate evaluation and Safety violation
rate for real {Laptop} experiments. Each task is
evaluated with 10 trials.}

\centering{}\resizebox{0.45\textwidth}{!}{
\begin{tabular}{c|c|c|c|c|c}
\hline 
{Laptop Task} & {DexArt} & {$\mathrm{DexArt_{aff}}$} & {QVPO} & {QSM} & \textbf{{DART}}\tabularnewline
\hline 
{Laptop (Position)} & {0} & {20} & {0} & {40} & \textbf{{80}}\tabularnewline
{Laptop (Orientation)} & {0} & {0} & {0} & {60} & \textbf{{90}}\tabularnewline
{Average (Mean)} & {0} & {10} & {0} & {50} & \textbf{{85}}\tabularnewline
\hline 
{Safety Violation Rate} & {DexArt} & {$\mathrm{DexArt_{aff}}$} & {QVPO} & {QSM} & \textbf{{DART}}\tabularnewline
\hline 
{Laptop (Position)} & {100} & {80} & {0} & {0} & \textbf{{0}}\tabularnewline
{Laptop (Orientation)} & {100} & {100} & {0} & {0} & \textbf{{0}}\tabularnewline
{Average (Mean)} & {100} & {90} & {0} & {0} & \textbf{{0}}\tabularnewline
\hline 
\end{tabular}}
\end{table}

\begin{figure}
\centering{}\includegraphics[scale=0.18]{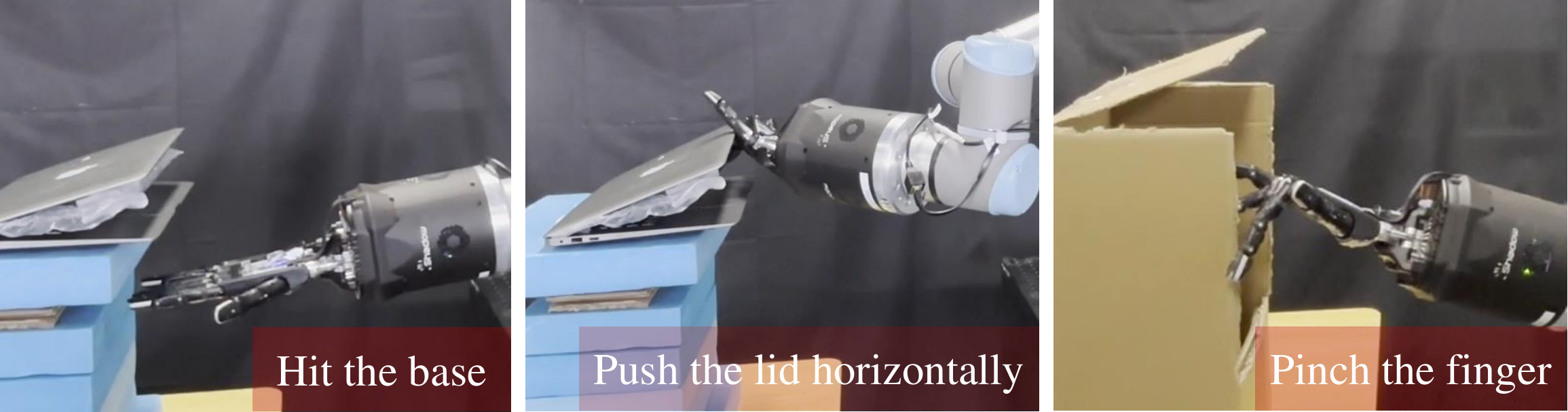}\caption{\label{fig:safety_violation}Example of safety violation when evaluating
DexArt and $\mathrm{DexArt_{aff}}$ for the real-world experiment.}
\end{figure}

{\textit{2) Safety violation:} In real-world experiments, we
observe that some methods may have unpredictable behaviors that must
be terminated to ensure safety. We define this situation as safety
violation and show the corresponding results in Table \ref{tab:real_evaluation}
and Fig. \ref{fig:safety_violation}. It can be observed that (1)
the operations of DexArt and $\mathrm{DexArt_{aff}}$ are highly likely
to cause damage to the robot due to the radical exploration of entropy;
(2) the methods based on policy characterization (QVPO, QSM, and DART)
although not necessarily having high success rates, exhibit actions
that are consistently safe, possibly due to the diffusion-based policy
performing $K$ iterations to gradually denoise a random action into
the noise-free action, which are not only robust to changes in the
environment but also synthesize multimodal features of action distributions
to avoid selecting rare or extreme, potentially dangerous actions.}


\section{CONCLUSIONS}


In this work, we introduce
DART, {which leverages LTL representations to enhance
task semantics understanding and improve the learning efficiency,
affordance learning to guide efficient manipulation of articulated
objects based on the predicted max-affordance point observation,} and {a
diffusion-based policy} to improve generalization across diverse object types. We enhance the action optimization method to address challenges inherent in traditional diffusion policies that rely solely on prior information. Extended experiments demonstrate the superior performance of DART over existing methods. 

Despite its great potential,
several challenges remain. First, when designing {the diffusion-based policy} in DART, we did not extend it to the classifier-guided case. This is because according to the experiments in \cite{pearce2023imitating} and \cite{reuss2023goal}, incorporating classifier-guided samples in the RL setting may degrade algorithm performance, even though it is a useful technique in computer vision. Another potential issue is that the Task Representation Module is updated indirectly through the gradient of the {diffusion-based policy}. Finding more ways to directly update the LTL encoder by utilizing rewards may further improve the performance of the algorithm. Hence, future research will consider {leveraging the policy with the classifier-guided
diffusion} to tackle more intricate dexterous manipulation tasks, such as tool understanding and utilization.

\bibliographystyle{IEEEtran}
\bibliography{Bib4DART}
\begin{IEEEbiography}[{\includegraphics[width=1in,height=1.25in,clip,keepaspectratio]{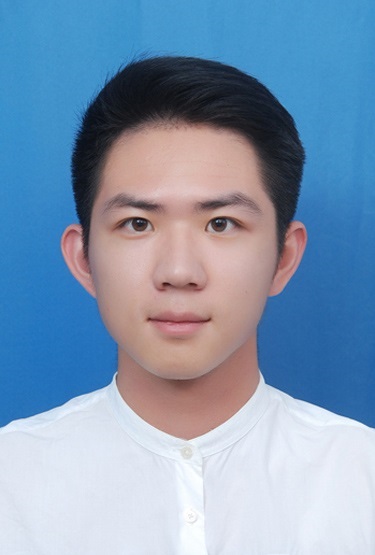}}]{Hao Zhang}
received the B.S. degree in mechanical engineering and automation from the Hefei
University of Technology, Hefei, Anhui, China, in
2020. He is currently pursuing the Ph.D. degree
in automation with the University of Science and
Technology of China, Hefei.

His current research interests include formal
methods in robotics, reinforcement learning, and
dexterous manipulation.
\end{IEEEbiography}

\begin{IEEEbiography}
[{\includegraphics[width=1in,height=1.25in,clip,keepaspectratio]{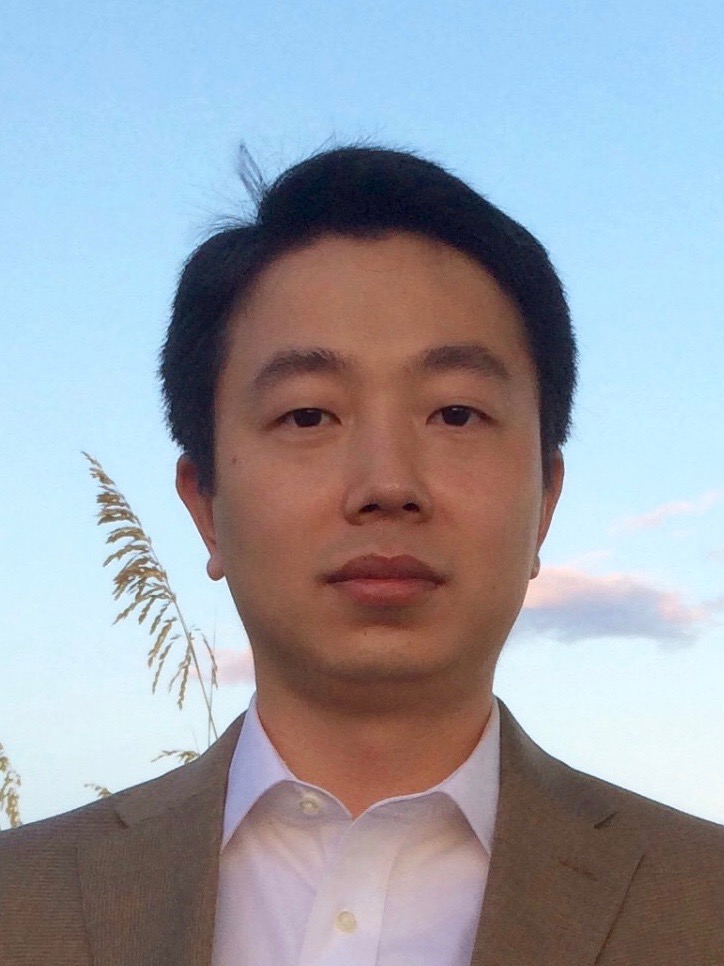}}]{Zhen Kan}
(Senior Member, IEEE) received the
Ph.D. degree from the Department of Mechanical
and Aerospace Engineering, University of Florida,
Gainesville, FL, USA, in 2011.

He is currently a Professor with the Department
of Automation, University of Science and
Technology of China, Hefei, China. His research interests include networked
control systems, nonlinear control, formal methods, and robotics. He currently serves on the program committees of several
internationally recognized scientific and engineering conferences and is an
Associate Editor of \textit{IEEE Transactions on Automatic Control}
\end{IEEEbiography}

\begin{IEEEbiography}
[{\includegraphics[width=1in,height=1.25in,clip,keepaspectratio]{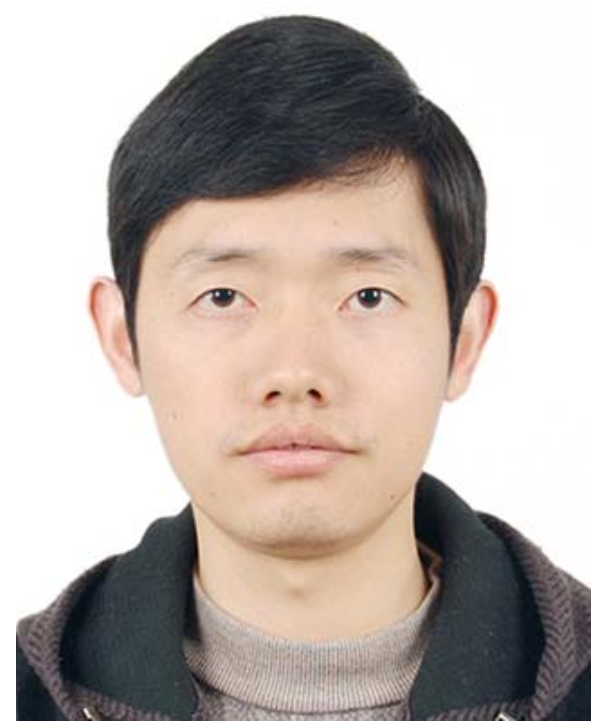}}]{Weiwei Shang}
(Senior Member, IEEE) received the Ph.D. degree in control science and engineering from the University of Science and Technology of China, Hefei, P.R. China, in 2008.

He is a Professor with the Department of Automation at the University of Science and Technology of China. His research interests
include parallel robots, cable-driven robots, humanoid robots, and robotic learning control.
\end{IEEEbiography}

\begin{IEEEbiography}
[{\includegraphics[width=1in,height=1.25in,clip,keepaspectratio]{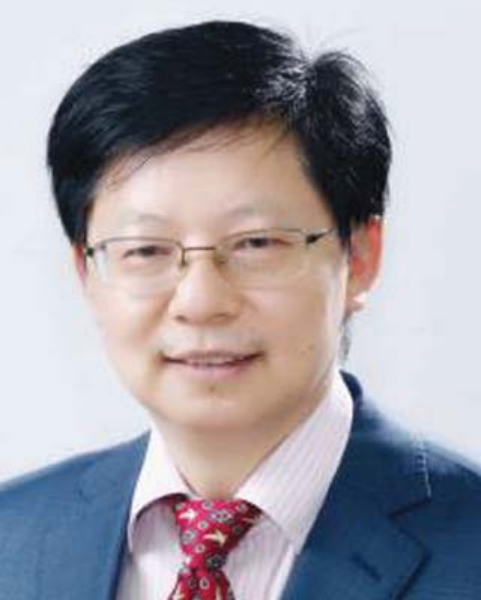}}]{Yongduan Song}
(Fellow, IEEE) received the Ph.D.
degree in electrical and computer engineering from
Tennessee Technological University, Cookeville,
TN, USA, in 1992.

Prof. Song is currently the Dean of the Research Institute
of Artificial Intelligence, Chongqing University,
Chongqing, China. He is the Editor-in-Chief of
\textit{IEEE Transactions on Neural Networks and Learning Systems} and the Founding
Editor-in-Chief of the \textit{International Journal of
Automation and Intelligence}. He was one of the six Langley Distinguished
Professors at the National Institute of Aerospace, USA and a Register
Professional Engineer (USA). He is a Fellow of AAIA, the International
Eurasian Academy of Sciences, and the Chinese Automation Association.
\end{IEEEbiography}

\end{document}